%% file: neurips_2025.tex
\documentclass{article}

\usepackage[preprint]{neurips_2025}

\usepackage[utf8]{inputenc} 
\usepackage[T1]{fontenc}    
\usepackage{hyperref}       
\usepackage{url}            
\usepackage{booktabs}       
\usepackage{tabularx}   
\usepackage{siunitx} 
\usepackage{amsfonts}       
\usepackage{amsmath}       
\usepackage{nicefrac}       
\usepackage{microtype}      
\usepackage{xcolor}         
\usepackage{graphicx} 
\usepackage{subcaption}
\usepackage{epsfig}
\usepackage{array}     
\usepackage{multirow}
\usepackage{makecell} 
\usepackage{hhline}
\usepackage{enumitem}
\usepackage[dvipsnames]{xcolor}

\title{UniVerse-1: \underline{Uni}fied Audio-\underline{V}ideo Gen\underline{er}ation via \underline{S}titching of \underline{E}xperts}

\author{
    \textbf{Duomin Wang}$^{1}$ \quad 
    \textbf{Wei Zuo}$^{1}$ \quad 
    \textbf{Aojie Li}$^{1}$ \quad 
    \textbf{Ling-Hao Chen}$^{1,4}$ \quad
    \textbf{Xinyao Liao}$^{1}$ \quad  \\
    \textbf{Deyu Zhou}$^{1,2}$ \quad 
    \textbf{Zixin Yin}$^{1,3}$ \quad 
    \textbf{Xili Dai}$^{2}$ \quad 
    \textbf{Daxin Jiang}$^{1}$ \quad 
    \textbf{Gang Yu}$^{1}$ \quad 
    \\
    StepFun$^{1}$ \quad \\
    The Hong Kong University of Science and Technology(GuangZhou)$^{2}$ \quad \\
    The Hong Kong University of Science and Technology$^{3}$ Tsinghua University$^{4}$\quad \\
}

\begin{document}

\maketitle

\begin{abstract}
We introduce \textbf{UniVerse-1}, a unified, Veo-3-like model capable of simultaneously generating coordinated audio and video. To enhance training efficiency, we bypass training from scratch and instead employ a \textbf{stitching of experts (SoE)} technique. This approach deeply fuses the corresponding blocks of pre-trained video and music generation experts models, thereby fully leveraging their foundational capabilities. To ensure accurate annotations and temporal alignment for both ambient sounds and speech with video content, we developed an \textbf{online annotation pipeline} that processes the required training data and generates labels during training process. This strategy circumvents the performance degradation often caused by misalignment text-based annotations. Through the synergy of these techniques, our model, after being finetuned on approximately $7,600$ hours of audio-video data, produces results with well-coordinated audio-visuals for ambient sounds generation and strong alignment for speech generation. To systematically evaluate our proposed method, we introduce \textbf{Verse-Bench}, a new benchmark dataset. In an effort to advance research in audio-video generation and to close the performance gap with state-of-the-art models such as Veo3, we make our model and code publicly available. We hope this contribution will benefit the broader research community. Project page: \url{https://dorniwang.github.io/UniVerse-1/}.
\end{abstract}

\input{sections/1_Introduction}
\input{sections/2_RelatedWorks}
\input{sections/3_Methods}

\input{sections/4_Experiments}
\input{sections/5_Limitation}
\input{sections/6_Conclution}

\newpage

{
    \small
    \bibliographystyle{unsrt}
    \bibliography{neurips_2025}
}

\newpage
\input{sections/Appendix}

\end{document}

%% file: sections/1_Introduction.tex
\section{Introduction}

The era of diffusion models~\cite{song2020score, song2020improved, lipman2022flow} has culminated in the rise of Diffusion Transformer (DiT) architectures~\cite{peebles2023scalable}, exemplified by landmark models like Sora~\cite{openaisora2024} and its open-source counterparts~\cite{kong2024hunyuanvideo,yang2024cogvideox,wan2025wan}. These models, leveraging unprecedented scales of data and computation, have achieved remarkable quality and prompt alignment in video generation. This success is catalyzing a profound transformation across creative industries and has spurred a wave of research into downstream applications such as talking head synthesis~\cite{wang2023progressive,yu2023talking,wang2025fantasytalking,luo2025dreamactor} and human animation~\cite{tan2024animate,chen2025hunyuanvideo,lin2025omnihuman}, which now offer viable real-world solutions.

However, this rapid progress has been almost exclusively confined to the visual domain, treating video as a silent movie. This \textbf{unimodal focus} represents a fundamental bottleneck, as it ignores the inherently multimodal nature of video. Post-hoc video-to-audio models~\cite{shan2025hunyuanvideo} serve as a superficial fix, but they are inherently limited; while capable of adapting audio to existing visual content, they fail to enforce temporal alignment in the reverse direction. This makes critical tasks, such as synchronizing lip movements with speech, impossible. While closed-source systems like Google's Veo3~\cite{veo32025} have demonstrated synchronous audio-video generation, the lack of publicly available technical details leaves a critical gap in open research.

To bridge this gap between closed-source systems and open research, we introduce \textbf{UniVerse-1}: a unified, fully open-source, Veo-3-like model capable of simultaneously generating coordinated audio and video. Our work is underpinned by several key technical contributions designed to address the unique challenges of bimodal generation.

Instead of the costly process of training a new model from scratch, we propose a novel and efficient \textbf{stitching of experts (SoE)} paradigm. This methodology effectively fuses a state-of-the-art video generation model, WAN2.1~\cite{wan2025wan}, with a music generation model, Ace-step~\cite{gong2025ace}. The core of this fusion lies in lightweight, cross-modal MLP connectors introduced within corresponding blocks of each model. These connectors facilitate bidirectional interaction between modalities, and we found this strategy to significantly accelerate training convergence by leveraging the powerful priors of the pre-trained experts.

Furthermore, we tackle the critical challenge of data alignment in bimodal training. We argue that static, pre-processed annotations are a flawed paradigm for tasks requiring precise temporal consistency. To address this, we developed an \textbf{online annotation pipeline} that generates labels dynamically during training. This approach ensures strict temporal and semantic alignment between audio-video data and their textual descriptions, mitigating the performance degradation caused by static misalignment. During our investigation, we also uncovered a crucial, yet overlooked, factor in bimodal diffusion modeling: \textbf{cross-modal noise correlation}. We identified that the standard pseudo-random number generation process~\cite{hamming1952mathematical} can introduce spurious correlations between the noise vectors for video and audio, which subsequently degrades audio quality during inference. Our solution involves ensuring independent noise sampling for each modality.

To support this work, we curated a high-quality dataset comprising approximately $7,600$ hours of precisely aligned audio-video content. To systematically evaluate our method, we also propose \textbf{Verse-Bench}, a new benchmark featuring $600$ image-text prompt pairs covering a diverse range of sound categories. Highlighting its versatility, Verse-Bench supports not only joint audio-video generation but also unidirectional tasks, including a specialized \textbf{Verse-Ted} subset designed for evaluating audio-to-video synthesis.

In summary, our primary contributions are:
\begin{itemize}[leftmargin=*]
    \item \textbf{An Open-source Audio-Video Foundation Model:} We present UniVerse-1, a novel, open-source model capable of producing highly coherent and well-aligned synchronous audio-visual content, closing a critical gap in the open-source community.
    
    \item \textbf{A Novel Methodology for Joint Audio-Video Generation:} We propose a comprehensive methodology to enable efficient and high-quality joint audio-video synthesis. This is achieved through three key innovations: a \textbf{stitching of experts (SoE)} paradigm to accelerate convergence by fusing pre-trained models; an \textbf{online data annotation pipeline} to solve the critical static misalignment problem in training; and the identification and mitigation of the previously overlooked \textbf{cross-modal noise correlation} issue, a crucial factor for generation quality.
    
    \item \textbf{A Comprehensive Evaluation Benchmark:} We propose Verse-Bench, a new benchmark designed to comprehensively evaluate joint audio-video generation models across a diverse set of tasks.
\end{itemize}

%% file: sections/2_RelatedWorks.tex
\section{Related Works}
\paragraph{Video Diffusion Models} The field of video generation was revolutionized by the introduction of diffusion models, with pioneering works like AnimateDiff~\cite{guo2023animatediff} and Video Diffusion Models~\cite{ho2022video} marking the beginning of this new era. This initial wave of research was further advanced by models such as Stable Video Diffusion~\cite{blattmann2023stable}, which first demonstrated that curating large-scale, high-quality datasets is critical for enhancing model performance. A common characteristic of these early models was their reliance on UNet architectures. To mitigate the challenges posed by the limited availability and quality of video data compared to images, these models were typically fine-tuned from pre-trained UNet image foundations.
A significant paradigm shift occurred with the introduction of Sora~\cite{openaisora2024}, which heralded a new age defined by the Diffusion Transformer (DiT) architecture~\cite{peebles2023scalable} and training on massive, high-quality video corpora. This breakthrough spurred a proliferation of subsequent research. CogVideox~\cite{yang2024cogvideox} was the first to release an open-source DiT-based model, providing a significant catalyst for community-driven innovation. This was followed by other notable open-source models such as HunyuanVideo~\cite{kong2024hunyuanvideo}, WAN2.1~\cite{wan2025wan}, and Step-Video~\cite{ma2025step}, as well as high-performing closed-source systems including Kling~\cite{kuaishou2024kling}, SeeDance 1.0~\cite{gao2025seedance}, Movie Gen~\cite{polyak2024movie}, and Veo2~\cite{veo22025}.
Architecturally, these contemporary models converge on a common blueprint. They employ a 3D Variational Autoencoder (VAE) to achieve spatio-temporal compression of video into a latent space. The core generative process is then handled by a DiT, which learns to denoise these noisy latents. Across these state-of-the-art models, the quality and scale of the training data have been identified as paramount factors, making data curation and processing a central component of their development.

\paragraph{Audio Diffusion Models} The application of diffusion models to audio generation has followed a parallel trajectory to their video counterparts, fundamentally transforming the landscape of text-to-audio and text-to-music synthesis. Early explorations demonstrated the potential of diffusion for generating high-fidelity audio, but a pivotal advancement was the adoption of latent diffusion architectures~\cite{rombach2022high}, which significantly improved both efficiency and quality. A common technical pipeline for these approaches involves first transforming the raw audio waveform into a mel spectrogram. A Variational Autoencoder (VAE) is then trained on this spectrogram representation to learn a compressed latent space, within which the core diffusion process operates. Within this framework, models such as Stable Audio Open~\cite{evans2025stable} and Riffusion~\cite{riffusion} excel at generating high-fidelity, long-form audio. Furthermore, models like the AudioLDM series~\cite{audioldm2-2024taslp,liu2023audioldm} and DiffRhythm~\cite{ning2025diffrhythm} advance these capabilities to vocal music synthesis, offering fine-grained control over rhythm and other expressive attributes.

\paragraph{Joint Audio and Video Generation} The exploration of joint audio-video generation within diffusion frameworks began with pioneering efforts like MM-Diffusion~\cite{ruan2023mm}. This model was the first to tackle this bimodal task, employing a UNet architecture with two distinct subnetworks, each dedicated to processing the audio and video modalities, respectively. Following this initial work, a series of subsequent models emerged~\cite{ishii2024simple,hayakawa2024mmdisco,ergasti2025r}. However, these early approaches were typically constrained by small-scale training datasets~\cite{lee2022sound,li2021ai}, often less than $10$ hours in size, which inherently limited their diversity and generalization capabilities. A notable step forward was made by models such as Syncflow~\cite{liu2024syncflow} and Uniform~\cite{zhao2025uniform}, which scaled up the training data to approximately $500$ hours by leveraging the VGGSound~\cite{chen2020vggsound} and AudioSet~\cite{gemmeke2017audio} dataset, thereby enhancing their generalization. Despite this progress, persistent challenges remained, including suboptimal video quality and a low degree of disentanglement between the audio and visual streams. The advent of Google's Veo3~\cite{veo32025} marked a significant milestone, representing the first large-scale initiative in synchronous audio-video generation. Veo3 demonstrated the capacity for generating high-fidelity audio and video that is not only diverse but also semantically and temporally coordinated, strictly adhering to user-provided text prompts.

%% file: sections/3_Methods.tex
\section{UniVerse-1}

\subsection{Preliminary}
\label{sec:preliminary}


Our model is constructed upon the foundations of the Wan2.1 (1.3B parameters)~\cite{wan2025wan} text-to-video model and the Ace-step (3.5B parameters)~\cite{gong2025ace} music generation model. Before delving into technical details, we will briefly introduce their respective architectures.


\textbf{Wan2.1 model.} Wan2.1 is composed of three primary components: a 3D Variational Autoencoder (VAE), an umT5~\cite{chung2023unimax} text encoder, and a Diffusion Transformer (DiT). The 3D VAE compresses an input video of shape $(3, T, H, W)$ into a latent representation of shape $(16, T/t, H/h, W/w)$, where the temporal and spatial downsampling factors are $t=4$ and $h=w=8$, respectively. Prior to being input to the DiT, this latent tensor is patchified using a kernel of $(1, 2, 2)$, and the resulting tokens serve as the input sequence. The umT5 model encodes the text prompt, and its embeddings are injected into the DiT via cross-attention to condition the generation. The model is trained to predict the velocity, which is used in the denoising step to recover the clean latent. Finally, the decoder of 3D VAE reconstructs this latent into the final video.


\textbf{Ace-step model.} Ace-step consists of a Music-DCAE (Deep Compression Autoencoder)~\cite{chen2024deep}, a umT5 text encoder, a lyric encoder, a speaker encoder, and a DiT. The raw audio waveform is first converted into a mel spectrogram. The Music-DCAE then encodes the input spectrogram of shape $(8, T, F)$ into a latent representation of shape $(8, T/t, F/f)$, with downsampling factors $t=f=8$ along the temporal and frequency axes. This latent is subsequently patchified using a kernel of $(16, 1)$ to produce the input tokens for the DiT. Conditional control is provided by three sources: the umT5 encoder for the music style prompt, the lyric encoder for the lyrics, and the speaker encoder for the speaker ID. These three embeddings are concatenated along the channel dimension and injected into the DiT via cross-attention. The model predicts the velocity to obtain the clean latent, which the Music-DCAE's decoder reconstructs into a mel spectrogram. A HiFiGAN vocoder~\cite{liao2024fish} is then used to convert this spectrogram into the final audio waveform.


\textbf{Base model pre-training.} The learning objective for both of the aforementioned models is \textit{Flow Matching}~\cite{lipman2022flow}. This fashion trains a neural network, $v_\Theta(\cdot, t, c)$, to predict a velocity field that transports samples from a simple source distribution, $p_0$ (\textit{e.g.}, Gaussian noise), to a complex target data distribution, $p_1$. Specifically, these models leverage Conditional Flow Matching. Given a noise sample $x_0 \sim p_0$ and a data sample $x_1 \sim p_1$, a simple linear interpolation path is defined for time $t$:
\begin{equation}
    x_t = (1-t)x_0 + t x_1.
\end{equation}
The target velocity vector along this path is constant: $u_t = x_1 - x_0$. The model $v_\Theta(x_t, t, c)$, conditioned on $c$, is trained to predict this vector by minimizing the following L2 loss:
$$
\mathcal{L}_{\text{FM}} = \mathbb{E}_{t \sim U(0,1), x_0 \sim p_0, x_1 \sim p_1} \left[ \| v_\theta((1-t)x_0 + t x_1, t, c) - (x_1 - x_0) \|^2 \right].
$$
This objective directly trains the model to learn the vector field that maps noise to data, which leads to more stable and efficient training compared to traditional score-matching objectives.

\subsection{Data Curation}
We curated a large-scale, high-quality dataset from a diverse range of sources to train our model. The primary component was sourced from YouTube, encompassing content such as music variety shows, classical music performances, cooking tutorials, public speeches, interviews, vlogs, and demonstrations of tool usage. This was supplemented with cinematic movie clips and high-quality stock footage from Pexels. To further bolster the audio modality, we also incorporated the widely-used VGGSound and AudioSet datasets.

For our self-collected data (YouTube, Pexels, and movie clips), we implemented a rigorous multi-stage filtering pipeline to ensure data quality and relevance:
\begin{itemize}[leftmargin=*]
    \item \textbf{Audio-Visual Pre-screening} Videos lacking an audio track were immediately discarded.
    \item \textbf{Quality Control} We filtered out content based on technical specifications: resolution below $1080p$, a bitrate-to-resolution ratio under $600$, and a DOVER~\cite{wu2023exploring} aesthetic quality score below $0.6$.
    \item \textbf{Temporal Coherence} PySceneDetect\footnote{https://github.com/Breakthrough/PySceneDetect} was applied to segment videos, and any resulting clip shorter than $5$ seconds was removed to ensure meaningful duration.
    \item \textbf{Audio Activity Detection} To eliminate silent segments, we analyzed each audio track for metrics such as volume, energy, and zero-crossing rate.
    \item \textbf{Speech Content Verification} Whisper~\cite{radford2023robust} was used to detect the presence of human speech. Clips without speech were retained as general audio-visual data. If speech was present, it proceeded to the next step.
    \item \textbf{Human Face Detection} For clips identified as containing speech, a second verification step was performed: we detected for the presence of a human face~\cite{deng2020retinaface}. If no face was found, the clip was discarded. If a face was present, we employed SyncNet~\cite{chung2016out} to verify the audio-visual correspondence (lip-sync). Only clips with a SyncNet confidence score above a threshold of $2.0$ were retained and explicitly labeled as containing speech content.
\end{itemize}

For the VGGSound and AudioSet data, a simplified process was used where we either performed scene detection or segmented clips based on their existing timestamps, retaining only those longer than $5$ seconds.

Following this comprehensive curation process, our final dataset comprises $7,685$ hours of data. This is categorized into three subsets: $1,187$ hours of verified speech-centric content, $3,074$ hours of general-purpose audio-video data, and $3,422$ hours from VGGSound and AudioSet primarily used for bolstering audio-specific training.

\subsection{Online Data Annotation}
Conventional offline annotation methods, where captions are pre-generated for entire videos, present a significant challenge for training generative models. During training, fixed-length clips are randomly sampled from these videos, often creating a temporal and semantic misalignment between the sampled clip and the global, pre-existing caption. This issue is particularly acute in the context of joint audio-video generation, where the temporal synchronization between an acoustic event and its description is critical. Even minor temporal shifts can render an audio annotation invalid.

To overcome these limitations, we propose and implement an Online Data Annotation Pipeline. This pipeline operates as a dedicated server process that runs concurrently with training. It dynamically fetches raw video, processes clips in real-time to generate precisely aligned data-annotation pairs, and populates a shared buffer. The main training process then acts as a consumer, fetching these ready-to-use data tuples, ensuring that every training instance is perfectly synchronized.

The online processing for each data tuple involves the following steps:
\begin{itemize}[leftmargin=*]
    \item \textbf{Temporal Sampling} A fixed-length segment (e.g., $5$ seconds) is randomly extracted from a source video, yielding corresponding video and audio streams.
    \item \textbf{Multi-modal Annotation} The extracted audio-video clip is immediately passed to our annotation module for captioning.
    \item \textbf{Text and Video Encoding} The generated text prompts (for video, audio, and speech) are encoded. Concurrently, the video clip is encoded into a spatio-temporal latent representation using the 3D VAE.
    \item \textbf{Audio Encoding} The audio stream is converted to a mel spectrogram and subsequently encoded into a latent representation by the Music-DCAE~\cite{chen2024deep}.
\end{itemize}

The core of our pipeline is the multi-modal annotation step (Step $2$), which proceeds as follows:
\begin{itemize}[leftmargin=*]
    \item \textbf{Speech Transcription} Whisper~\cite{radford2023robust} is employed to perform Automatic Speech Recognition (ASR) on the sampled audio, yielding the raw speech content.
    \item \textbf{Structured Multimodal Captioning} We construct a structured prompt that incorporates the transcribed speech. This prompt, along with the audio and video streams of the clip, is fed into the QWen2.5-Omni~\cite{xu2025qwen2} multimodal model. QWen2.5-Omni is specifically instructed to output three distinct, aligned annotations for the clip: the verified speech content, a descriptive video caption, and a caption for the ambient audio.
\end{itemize}

This online, just-in-time process guarantees that every training instance consists of video and audio latents that are perfectly synchronized in time and semantically consistent with their corresponding textual annotations, thereby eliminating the data misalignment problem inherent in offline methods.

\begin{figure}[t!]
  \centering
  \includegraphics[width=\linewidth]{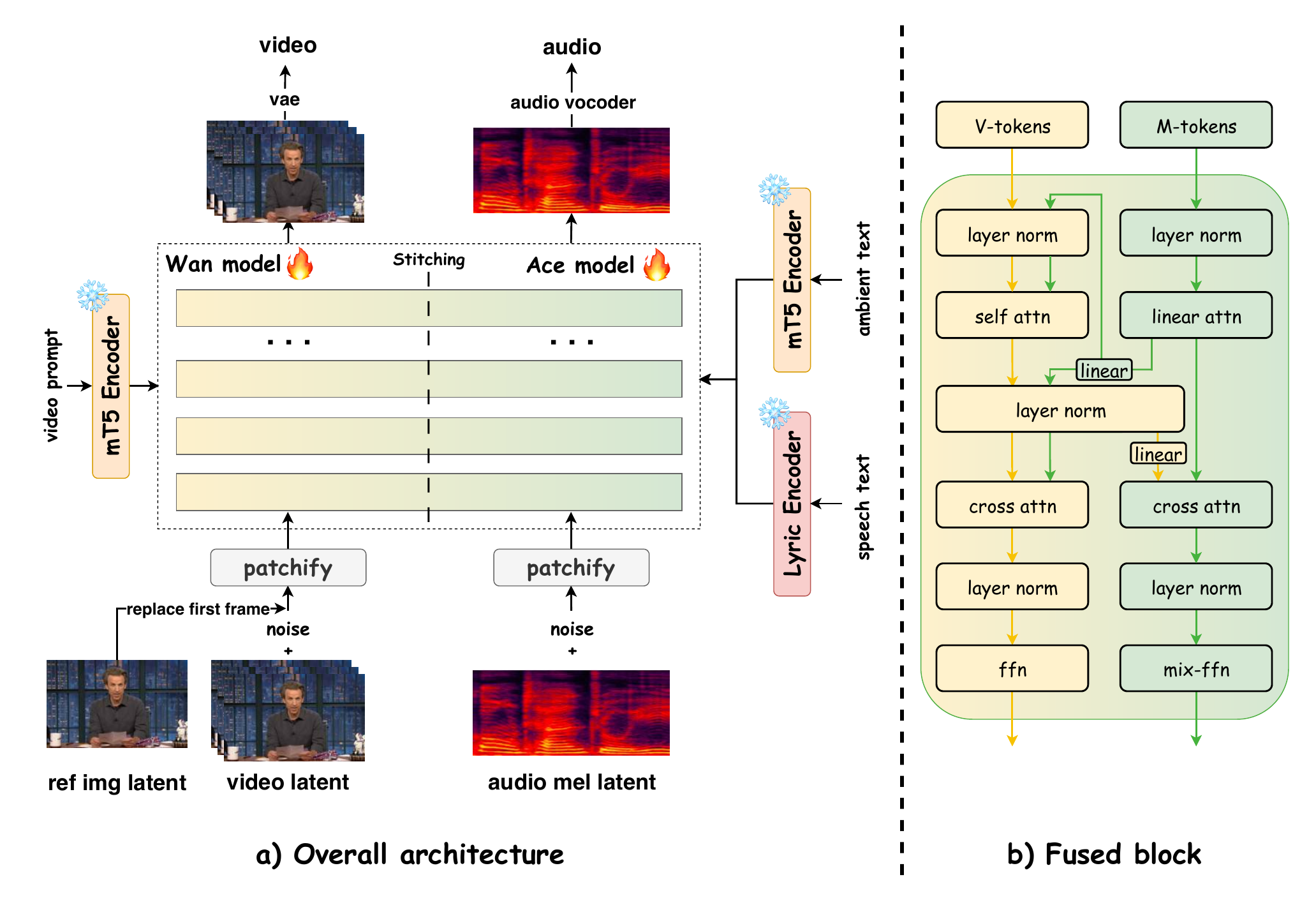}
  \caption{\textbf{Architecture of UniVerse-1.} (a) Overall architecture. The architectural foundation of UniVerse-1 is realized through a stitching of experts methodology. This approach deeply integrates the pre-trained Wan2.1 video model and the Ace-step audio model. (b) Fused block. The fusion is implemented at a granular, block-by-block level, where each block in the Wan architecture is deeply fused with its corresponding block in the Ace-step architecture.}
  \label{fig:archi}
\end{figure}

\begin{figure}[t!]
  \centering
  \includegraphics[width=\linewidth]{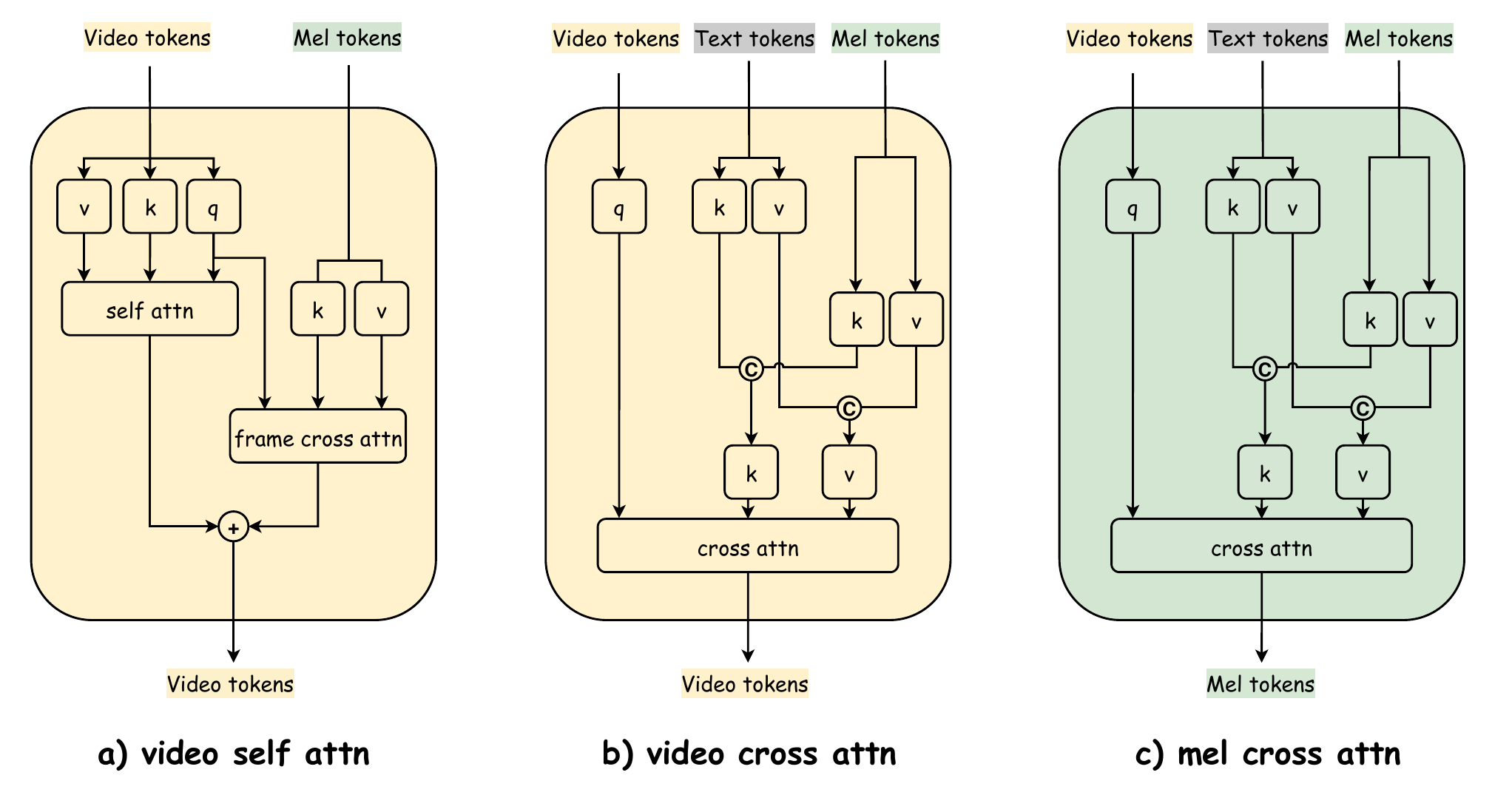}
  \caption{\textbf{Revised attention of UniVerse-1.} (a) Self attention of video branch, with additional mel tokens as input. (b) Cross attention of video branch, with additional mel tokens as input. (c) Cross attention of mel branch, with addition video tokens as input.}
  \label{fig:attn}
\end{figure}

\subsection{Method}
The overall architecture of our method is depicted in Fig.~\ref{fig:archi}. We introduce several targeted modifications to the input stages of the original Wan2.1 and Ace-step models to facilitate bimodal integration and control.
For the video component (Wan2.1), we enable conditioning on a reference image. During the forward process, the first frame of the noisy video latent is replaced with the corresponding clean latent representation of the provided reference image.
For the audio component (Ace-step), we perform two adjustments. First, to ensure temporal alignment with the video's 25 frames-per-second (fps) rate, the input mel spectrograms are processed at a 25.6 kHz sampling rate instead of the original 44.1 kHz. Second, to generalize the model beyond speaker-specific generation, we have removed the speaker encoder and its corresponding input from the architecture.

\subsubsection{Stitching of experts} We introduce a novel framework, termed ``Stitching of experts'', for integrating specialized, pre-existing models for video and audio synthesis. This approach is designed to preserve the generative capabilities of each unimodal expert while simultaneously enabling fine-grained, bidirectional interaction between them at the level of individual layer blocks. To enhance training efficiency and leverage the powerful priors of these pre-trained models, we apply the stitching technique to the Wan2.1 and Ace-step models at the transformer block level. This process results in a unified, dual-stream architecture where each block co-processes information from both the video and audio modalities, functioning akin to a Mixture-of-Experts (MoE) layer. \\
As illustrated in Fig~\ref{fig:archi}. b), we facilitate bidirectional cross-modal communication within each block. Specifically, the hidden states from the video stream, following its self-attention module, are injected into the audio stream's cross-attention module. Conversely, the hidden states from the audio stream, following its linear attention module, are reciprocally injected into the video stream's self-attention and cross-attention module. To ensure consistent feature scaling, the hidden states from both streams are jointly passed through a shared LayerNorm layer before injected into video stream's attention.\\
Prior to injection, the cross-modal hidden states are passed through a two-layer linear adapter for feature space alignment. In the video stream, for instance, features from the audio branch are projected using dedicated key ($k_{proj}$) and value ($v_{proj}$) layers~\ref{fig:attn}(a). A frame-by-frame cross-attention is then performed with the queries from the video stream to ensure alignment between the video and audio. Within each stream's respective cross-attention mechanism(shown in ~\ref{fig:attn}(b) and ~\ref{fig:attn}(c)), the conditioning signal from the other modality is projected using dedicated key ($k_{proj}$) and value ($v_{proj}$) layers. These new key-value pairs are then concatenated with the original text-derived key-value pairs along the context dimension, thereby enriching the conditioning information with cross-modal context.
\subsubsection{Layer Interpolation} A key challenge in stitching the Wan2.1 and Ace-step models is the architectural mismatch in their depth, as they possess a different number of transformer blocks. To reconcile this disparity, we introduce a \textbf{layer interpolation technique}.
This method involves first calculating the difference in the number of layers. We then strategically insert new blocks at uniform intervals into the shallower of the two models until their depths align. Crucially, the parameters for each new block are initialized by linearly interpolating the weights of its immediately adjacent (bracketing) layers. This initialization strategy effectively bridges the architectural gap while ensuring a smooth performance trajectory during training, thereby mitigating the risk of training instability and severe performance oscillations.
\subsubsection{Training Loss}
In addition to the primary Flow Matching objective (Sec.~\ref{sec:preliminary}), we incorporate two additional loss functions.
\paragraph{Semantic Alignment Loss}
For the audio modality, we employ a \textbf{Semantic Similarity Loss} ($\mathcal{L}_{\text{SSL}}$), a technique consistent with Ace-step, to enhance the semantic fidelity of the generated audio. This loss operates by aligning an intermediate feature representation, $h_{\text{audio}}$, extracted from the audio stream of our fused block at a specific layer $L$ ($L=12$ in our configuration). This internal representation is aligned against target features derived from two expert, pre-trained audio models:

\begin{itemize}[leftmargin=*]
    \item \textbf{MERT}(Music Encoder Representations from Transformers)~\cite{li2023mert}, which provides a general musical representation, $h_{\text{mert}}$, with a dimensionality of 1024 x $T_m$ (at a 75 Hz frame rate).
    \item \textbf{mHuBERT}(multilingual HuBERT)~\cite{boito2024mhubert}, which provides a speech-centric representation, $h_{\text{mHuBERT}}$, with a dimensionality of 768 x $T_h$ (at a 50 Hz frame rate).
\end{itemize}

To compute this loss, the intermediate feature $h_{\text{audio}}$ is first processed by two separate projection heads ($\pi_{\text{MERT}}$ and $\pi_{\text{mHuBERT}}$) and temporally interpolated to match the dimensionality and sequence length of $h_{\text{MERT}}$ and $h_{\text{mHuBERT}}$, respectively. The semantic similarity loss is then defined as the negative cosine similarity, encouraging the model's internal representations to align with those of the expert models:
$$
\mathcal{L}_{\text{SSL}} = \frac{1}{2} (\text{cosineSim}(h'_{\text{audio}}, h'_{\text{MERT}}) + \text{cosineSim}(h'_{\text{audio}}, h'_{\text{mHuBERT}}))
$$
where $h'$ denotes the temporally aligned representations.

\paragraph{Low Quality Data Loss Strategy} 

The AudioSet and VGGSound datasets, while offering rich auditory diversity, are characterized by low visual fidelity. To leverage their strong audio content without corrupting the video generation quality, we employ a conditional loss scheme.
Specifically, the Flow Matching loss for the video modality ($\mathcal{L}_{\text{FM-video}}$) is only computed for samples originating from these two datasets when the diffusion timestep \textit{t} exceeds an empirically determined threshold. We set this threshold to $\tau = 800$ (out of 1000 total timesteps). This strategy is predicated on the principle that at high noise levels (i.e., for $t > \tau$), the model learns to capture coarse, low-frequency features of the video, such as general motion and structure, which are less affected by the poor visual quality. By excluding the loss calculation at lower noise levels, we prevent the model from overfitting to the high-frequency visual artifacts and noise present in these datasets. The loss for video modality is:
$$
\mathcal{L}_{\text{FM-video}} = \left\{
\begin{array}{rcl}
\mathcal{L}_{\text{FM}}(x_1, x_0, t) & & if ~~ x_1 \in \Theta ~~ or ~~ (x_1 \in \zeta ~~ and ~~ t > 800)\\
0~~~~~~~~~ & & else
\end{array} \right.
$$
where $x_0 \sim p_0$ is noise sample, $x_1 \sim p_1$ is data sample, $\zeta$ is data subset include vggSound and audioset, $\Theta$ is data subset exclude vggSound and audioset. The final training objective is a weighted sum of the flow matching and semantic alignment losses:
$$
\mathcal{L} = \mathcal{L}_{\text{FM-video}} + \mathcal{L}_{\text{FM-mel}} + \lambda_{SSL}\cdot\mathcal{L}_{\text{SSL}}
$$
where $\mathcal{L}_{\text{SSL}}$ is a hyperparameter controlling the influence of the SSL guidance, empirically set to $1.0$ according to Ace-step.

\subsection{Independent Noise Sampling Strategy}

Our empirical investigation reveals a critical sensitivity of multi-modal diffusion models to the pseudo-random number generation process. When a single, fixed random seed is used to initialize a training run, the noise tensors for the video (${\epsilon}_v$) and audio (${\epsilon}_a$) modalities are sampled sequentially from the same deterministic PRNG sequence. Due to the deterministic nature of the underlying algorithm (Linear Congruential method), this sequential sampling introduces a spurious structural correlation between the two noise tensors, which can be expressed as ${\epsilon}_a = f({\epsilon}_v)$. This violates the critical assumption that the noise vectors are statistically independent.

The model inadvertently learns this spurious correlation as a shortcut during training. Consequently, during inference, any alteration to the sampling of ${\epsilon}_v$, such as a change in video resolution or duration, propagates through the PRNG's state and alters the structure of the subsequently sampled ${\epsilon}_a$. This mismatch with the learned correlation results in a significant degradation of the audio generation quality.

To address this, we propose an \textbf{Independent Noise Sampling Strategy}. This approach isolates the noise generation for each modality by employing separate and independently seeded PRNG instances. This method effectively breaks the deterministic correlation, ensuring the noise vectors are statistically independent. As a result, the model becomes robust to variations in inference-time conditions, mitigating the issue of performance degradation.

%% file: sections/4_Experiments.tex
\section{Experiments}
\subsection{Setup}
\paragraph{Implementation Details}
The training is conducted with an effective batch size of $128$ over $50k$ steps on a $7,600$-hour audio-visual datasets built by our data curation pipeline,
using the AdamW optimizer with a learning rate of $5e-6$. We employ Fully Sharded Data Parallel (FSDP) for distributed training across multiple nodes, with a gradient accumulation step of $4$.
\paragraph{Compared Methods} We conduct a comprehensive evaluation of our model by benchmarking it against a suite of state-of-the-art baselines across several distinct generation tasks:
\begin{itemize}[leftmargin=*]
    \item Video Generation: We compare our model against leading text-to-video systems, including Wan2.2 (14B)~\cite{wan2025wan}, HunyuanVideo~\cite{kong2024hunyuanvideo}, CogVideoX-1.5, (5B)~\cite{yang2024cogvideox}, Kling 2.1~\cite{kuaishou2024kling}, and SeeDance 1.0~\cite{gao2025seedance}.
    \item Audio Generation: For the audio modality, we benchmark against established text-to-audio models such as Stable Audio Open~\cite{evans2025stable} and AudioLDM2~\cite{audioldm2-2024taslp}.
    \item Text-to-Speech (TTS): As a supplementary evaluation of vocal synthesis, we compare our model's performance against specialized TTS models, including CosyVoice~\cite{du2024cosyvoice}, CosyVoice2~\cite{du2024cosyvoice2}, and VibeVoice~\cite{peng2025vibevoice}.
    \item Audio-to-Video methods: We also compare out model with talking-based audio to video methods such as FantastyTalking~\cite{wang2025fantasytalking} and Wan-S2V~\cite{gao2025wan}.
    \item Video-to-Audio methods: As a closely related task to joint audio-visual generation, we also benchmarked our model on the video-to-audio (V2A) task, drawing comparisons with state-of-the-art methods such as HunyuanVideo-Foley~\cite{shan2025hunyuanvideo}.
    \item Joint Audio-Video Generation: For our core task of synchronous audio-visual synthesis, we compare our method against existing prompt-based joint generation models: SVG~\cite{ishii2024simple}. Since methods such as MM-Diffusion~\cite{ruan2023mm} and R-FLAV~\cite{ergasti2025r} are class-conditional generative models, a direct comparison with our approach is not applicable, we only compare with SVG in our report.
\end{itemize}
It is important to note that our model is constructed by stitching the pre-trained WAN2.1 (1.3B) and Step-Ace (3.5B) models. Consequently, the comparisons against the aforementioned state-of-the-art baselines are intended to provide a qualitative reference and situate our work, rather than to make a direct claim of superior performance.

\begin{figure*}
    \centering
    \begin{subfigure}[h]{0.3\textwidth}
    \centering
    \includegraphics[width=1.0\linewidth]{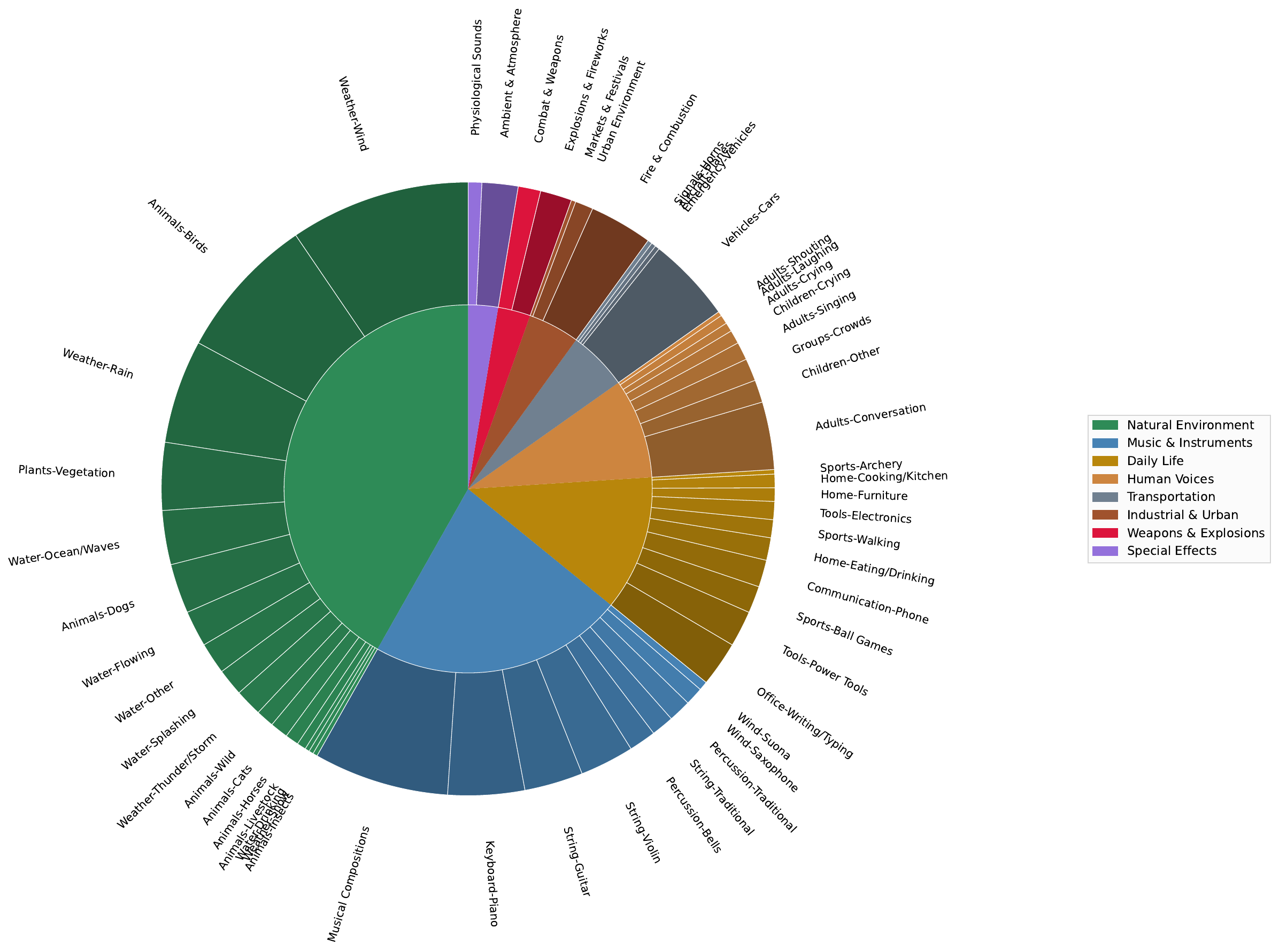}
    \caption{Statistical results of set1/2.}
    \label{fig:unibench_statistics_all}
    \end{subfigure}
    \hspace{0.01\textwidth} 
    \vrule
    \hspace{0.01\textwidth} 
    \begin{subfigure}[h]{0.3\textwidth}
    \centering
    \includegraphics[width=1.0\linewidth]{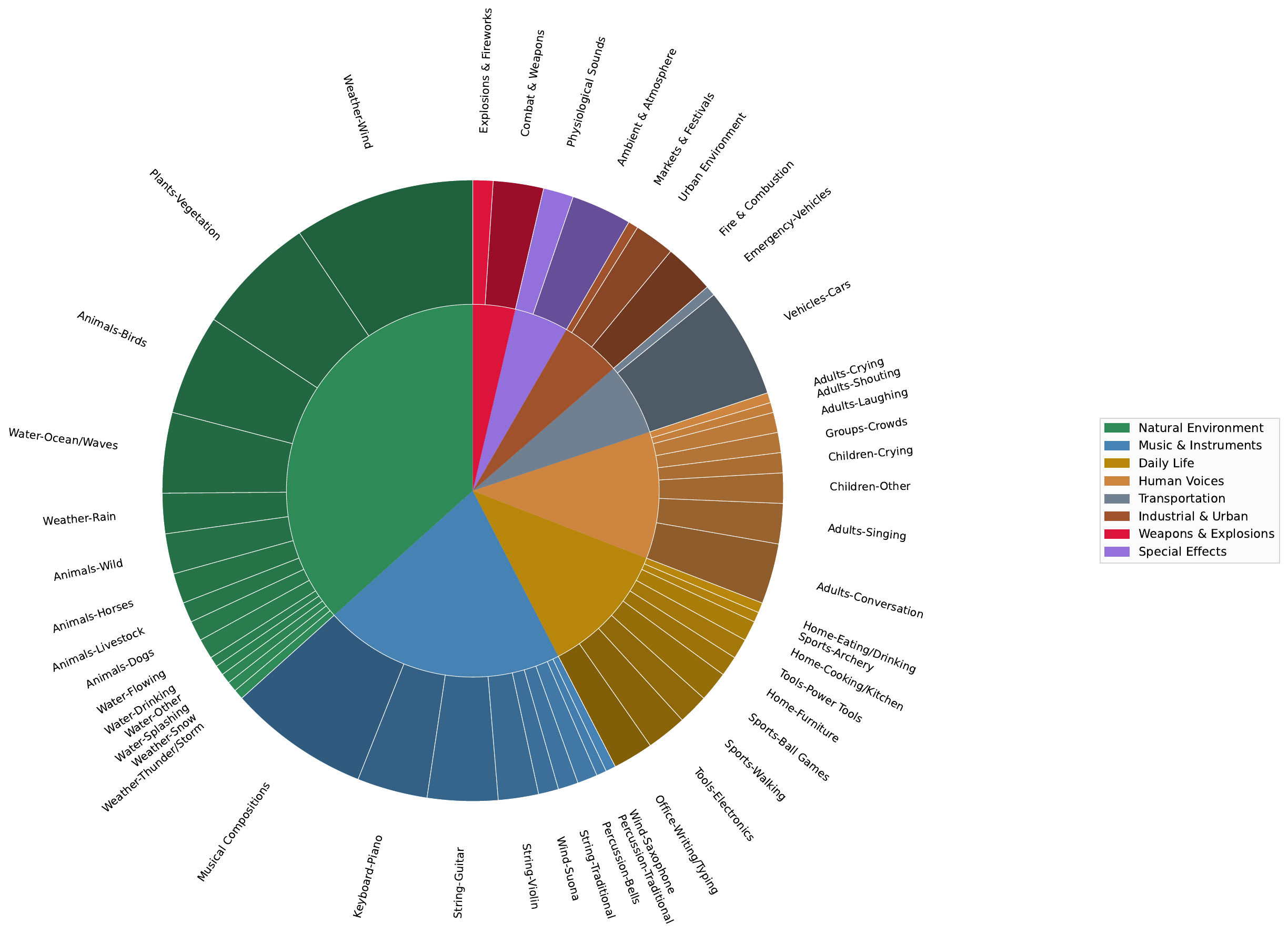}
    \caption{Statistical results of set1.}
    \label{fig:unibench_statistics_set1}
    \end{subfigure}
    \hspace{0.01\textwidth} 
    \vrule
    \hspace{0.01\textwidth} 
    \begin{subfigure}[h]{0.3\textwidth}
    \centering
    \includegraphics[width=1.0\textwidth]{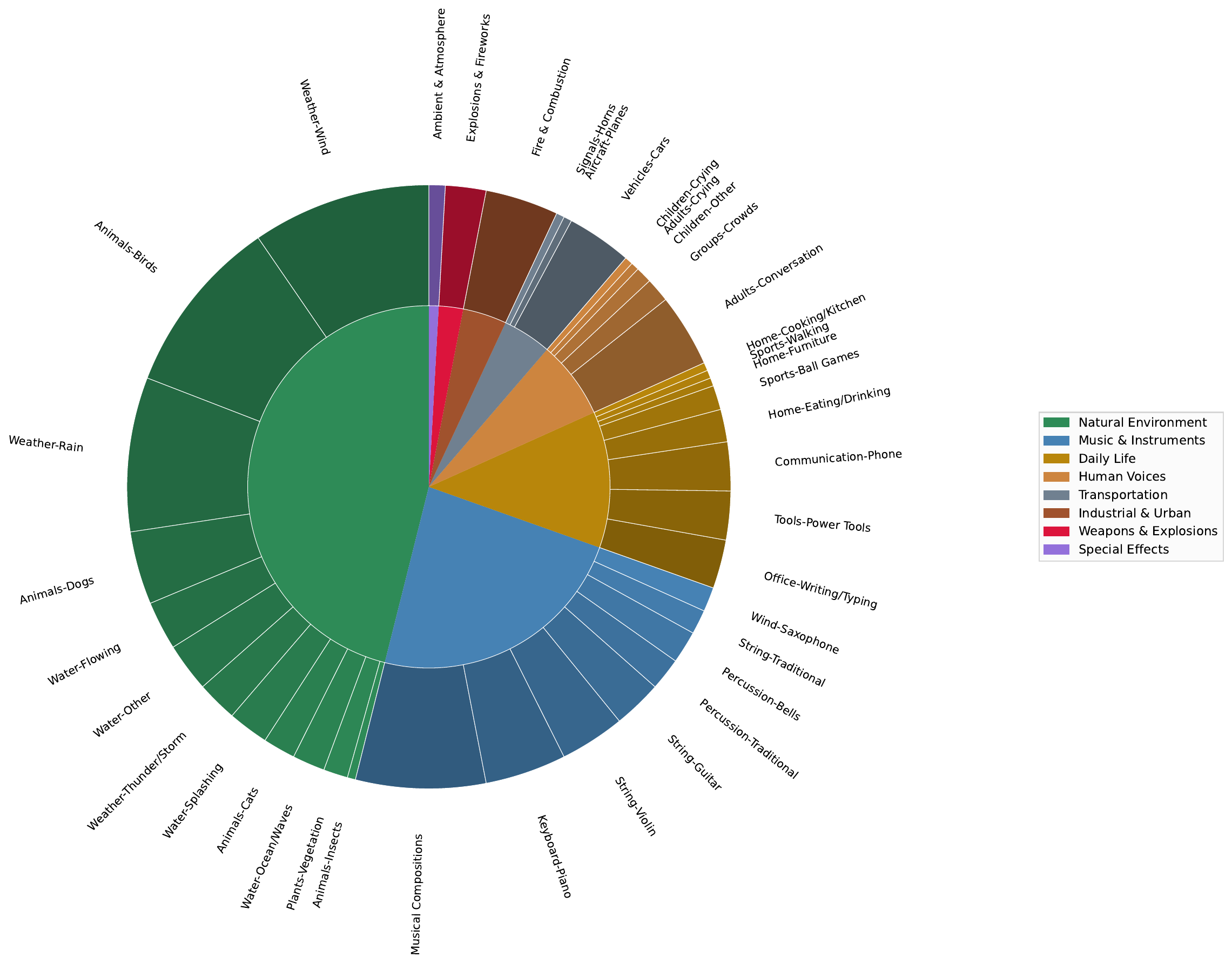}
    \caption{Statistical results of set2.}
    \label{fig:unibench_statistics_set2}
    \end{subfigure}
    
    \caption{Statistical results of \textbf{Verse-Bench}. Best viewed with \textbf{zoom-in}. A larger, high-resolution version is available in Appendix~\ref{lab:verse-bench-fig}}
\end{figure*}

\paragraph{Benchmark} To construct our evaluation set, we curated $600$ image-text prompt pairs from a multitude of sources. These sources encompass frames extracted from YouTube videos, BiliBili videos, TikTok clips, movies, and anime; images generated by AI models~\cite{jimeng,wu2025qwen}; and a collection of images from public websites. Our dataset comprises three subsets.
\begin{itemize}[leftmargin=*]
    \item Set1-I contains image-text pairs (including AI-generated, web-crawled, and media screenshots), for which video/audio captions and speech content were produced using LLMs~\cite{xu2025qwen2} and manual annotation, comprising a total of $205$ samples. Statistical results in~\ref{fig:unibench_statistics_set1}.
    \item Set2-V consists of video clips from YouTube and Bilibili, which were annotated with LLM-generated~\cite{xu2025qwen2} captions and Whisper-based ASR~\cite{radford2023robust} transcripts, followed by human verification, comprising a total of $295$ samples. Statistical results in~\ref{fig:unibench_statistics_set2}.
    \item Set3-Ted (the Verse-Ted subset) includes TED Talks from September 2025, processed with the same annotation pipeline as Set2, comprising a total of $100$ samples. 
\end{itemize}
Our collected test data is highly diverse, encompassing a wide spectrum of audio categories: human speech; animal vocalizations (e.g., bird chirping, cat meowing); instrumental music (e.g., piano, guitar); natural sounds (e.g., thunder, rain); human-object interactions (e.g., keyboard typing, cooking, chopping vegetables); object-object interactions (e.g., glass shattering, marbles dropping); mechanical sounds (e.g., trains, airplanes), and so on. The complete statistical results of set1 and set2 are shown in~\ref{fig:unibench_statistics_all}.

\paragraph{Evaluation Protocol}
We quantitatively evaluate our method against baselines on the Verse-Bench benchmark. The evaluation is structured across $6$ distinct generation tasks, each with a tailored set of metrics.

\begin{itemize}[leftmargin=*]
    \item Video Generation: Performance is assessed on three criteria:
    \begin{itemize}
        \item Motion Score (MS): This metric quantifies motion dynamics, calculated from the normalized optical flow magnitude detected by the RAFT model~\cite{teed2020raft}.
        \item Aesthetic Score (AS): This is a composite score averaging three components: fidelity, measured by MANIQA~\cite{yang2022maniqa} to penalize blur and artifacts, and aesthetic quality, evaluated by both aesthetic-predictor-v2-5~\cite{aesthetic-predictor-v2-5} and Musiq~\cite{ke2021musiq}.
        \item ID Consistency (ID): To measure identity preservation, we compute the mean DINOV3~\cite{simeoni2025dinov3} feature similarity between the reference image and each generated frame.
    \end{itemize} 
        
    \item Audio Generation: We evaluate audio quality from three perspectives:
    \begin{itemize}
        \item Distributional Similarity: We measure the Fréchet Distance (FD) and Kullback-Leibler (KL) divergence between the generated and real data distributions, using features extracted from PANNs~\cite{kong2020panns} and PaSST~\cite{koutini2021efficient}.
        \item Semantic Consistency: The alignment between the audio and the input text is measured by the LAION-CLAP~\cite{wu2023large} score.
        \item Quality and Diversity: We report the Inception Score (IS) calculated with a PANNs classifier. Additionally, we use AudioBox-Aesthetics~\cite{tjandra2025meta} to assess Production Quality (PQ), Production Complexity (PC), Content Enjoyment (CE), and Content Usefulness (CU).
    \end{itemize}
    \item Text-to-Speech (TTS): We evaluate synthesis accuracy using the Word Error Rate (WER), which is derived by transcribing the generated audio with the Whisper-large-v3 model~\cite{radford2023robust}.
    \item Audio-to-Video: We evaluate this task using the same criteria as the video generation task, additionally providing a SyncNet~\cite{chung2016out} confidence score to assess lip-sync accuracy.
    \item Video-to-Audio: This task use all metrics from audio generation tasks. Furthermore, we introduce the Audio-Video Alignment (AV-A) metric to specifically quantify the temporal synchronization between the generated audio and video streams, which is computed via Synchformer~\cite{iashin2024synchformer}.
    \item Joint Audio-Video Generation: For this task, we use all relevant metrics from the individual tasks above. 
\end{itemize}

The evaluation of our models and their components is conducted across the three test sets as follows:
\begin{itemize}[leftmargin=*]
    \item The video generation models and SVG are evaluated on Set 1 and Set 2.
    \item For the audio generation model is evaluated on Set1 and Set2.
    \item The Text-to-Speech (TTS) model is primarily evaluated on Set 3.
    \item The Audio-to-Video (A2V) model is evaluated exclusively on Set 3, while the Video-to-Audio (V2A) model is evaluated exclusively on Set 2.
    \item Finally, our complete Universe-1 model is benchmarked against all three test sets.
\end{itemize}
For audio generation, we evaluate the metrics CE, CU, PC, and PQ on Set 1. On Set 2, the evaluation is expanded to include FD, KL, and CS in addition to the aforementioned metrics. Furthermore, LSE-C is evaluated exclusively on Set 3, while the AV-A metric is also applied to Set 1 when evaluating UniVerse-1 and SVG.

\subsection{Quantitative Evaluation}

We compare Universe-1 against a range of state-of-the-art (SOTA) models specialized for specific tasks as shown in Tab.~\ref{tab:uni_bench}. As a unified joint generation model, a direct comparison of single-modality metrics against these "expert" models presents inherent complexities. Nevertheless, our model demonstrates robust capabilities across multiple dimensions.
In terms of video quality, Universe-1 achieves the highest score in identity preservation (ID: 0.89), showcasing its superior ability to maintain subject consistency throughout the generation process. For audio quality, while there is a gap compared to leading audio-only generation models, our model obtains a highly competitive score in pitch correlation (PC: 2.49).
The core strength of our model lies in synchronous audio-video generation. It is crucial to interpret the metrics for such joint generation tasks with caution. For instance, in the video-to-audio (V2A) setting, the AV-A metric for a model like Hunyuanvideo-Foley is calculated with a ground-truth video, whereas our model generates both modalities simultaneously. Therefore, AV-A must be considered in conjunction with the audio-text CLAP score (CS) for a holistic assessment. From this integrated perspective, our model (AV-A: 0.23, CS: 0.16) demonstrates a better overall audio-visual content consistency than SVG (AV-A: 0.09, CS: 0.08).
Similarly, for the lip-sync (LSE-C) metric, our model's score (1.34) is evaluated on fully generated audio and video, making it susceptible to the quality of both generated modalities. In contrast, audio-to-video (A2V) methods like Wan-S2V are evaluated using ground-truth audio, which naturally yields a higher score (6.49). Despite this evaluation disparity, Universe-1 achieves promising results as the first open-source joint generation framework of its kind, establishing a solid foundation for future research.

\begin{table}
\vspace{-1em}
\small
\centering
\resizebox{\linewidth}{!}{
    \renewcommand{\arraystretch}{1.8}
    \begin{tabular}{l|c|c|c|c|c|c|c|c|c|c|c|c|c|c}
    \toprule
    \multicolumn{2}{c|}{} & \multicolumn{3}{c|}{video} & \multicolumn{7}{c|}{audio} & \multicolumn{1}{c|}{tts} & \multicolumn{2}{c}{Audio-Video} \\
    \midrule
     & method & MS$~\uparrow$ & AS$~\uparrow$ & ID$~\uparrow$ & FD$~\downarrow$ & KL$~\downarrow$ & CS$~\uparrow$ & CE$~\uparrow$ & CU$~\uparrow$ & PC$~\downarrow$ & PQ$~\uparrow$ & WER$~\downarrow$ & LSE-C$~\uparrow$ & AV-A$~\downarrow$ \\
    \midrule
    
    \multirow{5}{*}{\makecell{video \\ methods}} 
     & CogVideox1.5 5B~\cite{yang2024cogvideox}         & 0.43 & 0.44 & 0.83 & - & - & - & - & - & - & - & - & - & - \\ \cline{2-15} 
     & Wan2.2-14B~\cite{wan2025wan}                     & \textbf{1.04} & \textbf{0.50} & \underline{0.88} & - & - & - & - & - & - & - & - & - & - \\ \cline{2-15} 
     & Kling2.1~\cite{kuaishou2024kling}                & 0.31 & 0.41 & 0.85 & - & - & - & - & - & - & - & - & - & - \\ \cline{2-15}
     & SeeDance1.0~\cite{gao2025seedance}               & \underline{0.50} & \underline{0.47} & 0.86 & - & - & - & - & - & - & - & - & - & - \\
    \midrule
    \multirow{2}{*}{\makecell{audio \\ methods}} 
     & Stable Audio~\cite{evans2025stable}              & - & - & - & \underline{1.13} & \underline{1.38} & \underline{0.28} & 3.88 & \textbf{6.15} & \underline{2.60} & 6.53 & - & - & - \\ \cline{2-15} 
     & AudioLdm2~\cite{audioldm2-2024taslp}             & - & - & - & 1.21 & 2.30 & 0.24 & \underline{3.98} & \underline{5.88} & 3.43 & 6.04 & - & - & - \\
    \midrule
    \multirow{3}{*}{\makecell{tts \\ methods}}   
     & Cosyvoice~\cite{du2024cosyvoice}                 & - & - & - & - & - & - & - & - & - & - & 0.17 & - & - \\ \cline{2-15}  
     & Cosyvoice2~\cite{du2024cosyvoice2}               & - & - & - & - & - & - & - & - & - & - & \underline{0.16} & - & - \\ \cline{2-15}
     & Vibevoice~\cite{peng2025vibevoice}               & - & - & - & - & - & - & - & - & - & - & \textbf{0.15} & - & - \\
    \midrule
    \multirow{2}{*}{\makecell{a2v \\ methods}}   
     & FantastyTalking~\cite{wang2025fantasytalking}    & 0.07 & 0.42 & 0.87 & - & - & - & - & - & - & - & - & 2.68 & - \\ \cline{2-15}  
     & Wan-S2V~\cite{gao2025wan}                        & 0.17 & 0.46 & \textbf{0.89} & - & - & - & - & - & - & - & - & 6.49 & - \\
    \midrule
    \multirow{1}{*}{\makecell{v2a \\ methods}}     
     & Hunyuanvideo-Foley~\cite{shan2025hunyuanvideo}   & - & - & - & \textbf{0.82} & \textbf{1.27} & \textbf{0.40} & \textbf{4.04} & 5.72 & 3.09 & 6.27 & - & - & 0.78 \\
    \midrule
    \multirow{2}{*}{\makecell{joint \\ methods}}  
     & SVG~\cite{ishii2024simple}                       & 0.40 & 0.41 & 0.25 & 1.55 & 3.62 & 0.08 & 2.93 & 5.50 & \textbf{2.35} & 6.26 & - & - & 0.09 \\ \cline{2-15} 
     & \colorbox{CornflowerBlue}{\textbf{Ours}}         & 0.20 & 0.47 & 0.89 & 1.25 & 2.70 & 0.16 & 3.53 & 4.61 & 2.49 & 5.20 & 0.18 & 1.34 & 0.23 \\ 
    \bottomrule
    \end{tabular}
}
\vspace{1em}
\caption{Quantitative results compared with baselines on Verse-Bench.}
\label{tab:uni_bench}
\vspace{-1em}
\end{table}

\subsection{Ablation Study}
We performed an ablation study to investigate the contributions of our Low Quality data Loss Strategy(LQLS) and Independent Noise Sampling Strategy(INSS), as shown in Tab.~\ref{tab:ablation}. The findings indicate that LQLS provides improvement accross video quality and consistency ID, thus confirming its efficacy and positive impact on training. Furthermore, the results for INSS demonstrate a significant enhancement in audio generation quality, validating the effectiveness of this approach.

\begin{table}
\small
\centering
\resizebox{\linewidth}{!}{
    \renewcommand{\arraystretch}{1.8}
    \begin{tabular}{l|c|c|c|c|c|c|c|c|c|c|c|c|c}
    \toprule
    method & MS$~\uparrow$ & AS$~\uparrow$ & ID$~\uparrow$ & FD$~\downarrow$ & KL$~\downarrow$ & CS$~\uparrow$ & CE$~\uparrow$ & CU$~\uparrow$ & PC$~\downarrow$ & PQ$~\uparrow$ & WER$~\downarrow$ & LSE-C$~\uparrow$ & AV-A$~\downarrow$ \\
    \midrule
     w/o LQLS                                  & 0.38 & 0.44 & 0.78 & 1.26 & 2.84 & 0.15 & 3.38 & 4.35 & 2.58 & 4.97 & 0.16 & 1.35 & 0.28 \\ \cline{1-14}
     w/o INSS                                  & 1.10 & 0.43 & 0.75 & 1.43 & 3.51 & 0.11 & 2.44 & 3.14 & 2.92 & 3.99 & 0.38 & 0.99 & 0.18 \\ \cline{1-14}
     \colorbox{CornflowerBlue}{\textbf{Ours}}  & 0.20 & 0.47 & 0.89 & 1.25 & 2.70 & 0.16 & 3.53 & 4.61 & 2.49 & 5.20 & 0.18 & 1.34 & 0.23 \\
    \bottomrule
    \end{tabular}
}
\vspace{1em}
\caption{Ablation results on Verse-Bench.}
\label{tab:ablation}
\vspace{-1em}
\end{table}

%% file: sections/5_Limitation.tex
\section{Limitation and Future Work}
The work presented herein constitutes an initial exploration into unified audio-video generation. Our study was constrained by computational resources, necessitating that we conduct training exclusively on the Wan2.1-1.3B video model. The performance of our model is, therefore, inherently limited by the capacity of this base model.
In the future, our research will focus on two key directions. First, we will scale up our experiments to larger video foundation models. Second, we will engage in more extensive and refined data curation efforts. The ultimate objective is to significantly advance the capabilities of open-source audio-video synthesis models, thereby bridging the performance gap to state-of-the-art proprietary models.

%% file: sections/6_Conclution.tex
\section{Conclusion}

In this paper, we presented UniVerse-1, a novel framework for joint audio-video synthesis, achieved through the deep integration of a video foundation model and a music generation model using stitching of experts. Following fine-tuning on the dataset we curated, we also introduced Verse-Bench, a comprehensive benchmark to foster comparative research. To promote reproducibility and further innovation, our model and code have been made publicly available. Our experimental results validate that this methodology offers a viable and efficient pathway for building sophisticated multimodal generative models by leveraging pre-existing unimodal foundations.

%% file: sections/Appendix.tex
\appendix
\section*{Appendix}
\addcontentsline{toc}{section}{Appendix}
\section{More Details about Verse-Bench}
\subsection{Statistical Results}
\label{lab:verse-bench-fig}
\begin{figure*}[h!]
    \centering
    \includegraphics[width=1.0\linewidth]{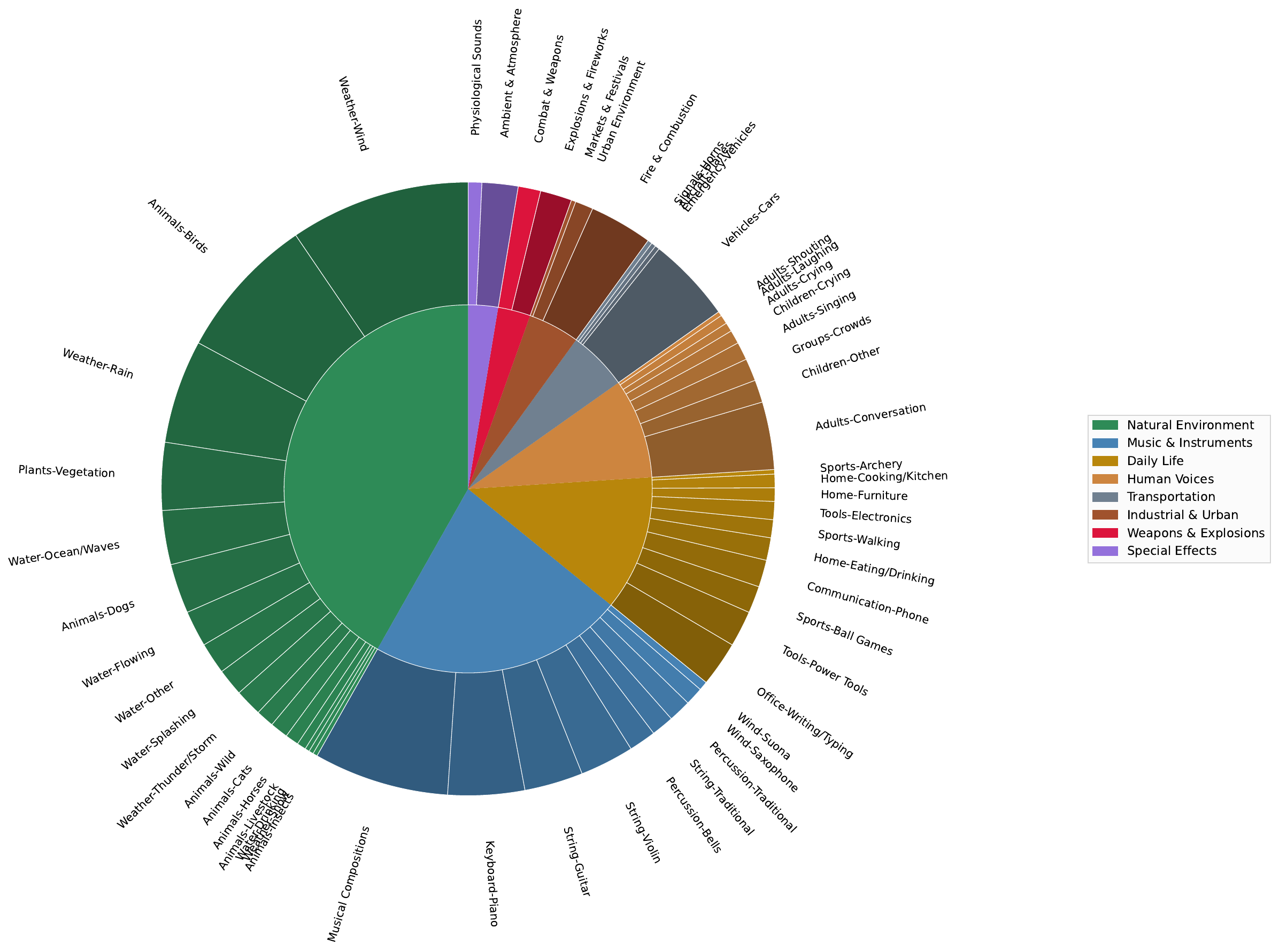}
    \caption{Statistical results of set1 and 2.}
    \label{fig:appendix_unibench_statistics_all}
\end{figure*}

\begin{figure*}[h!]
    \centering
    \includegraphics[width=1.0\linewidth]{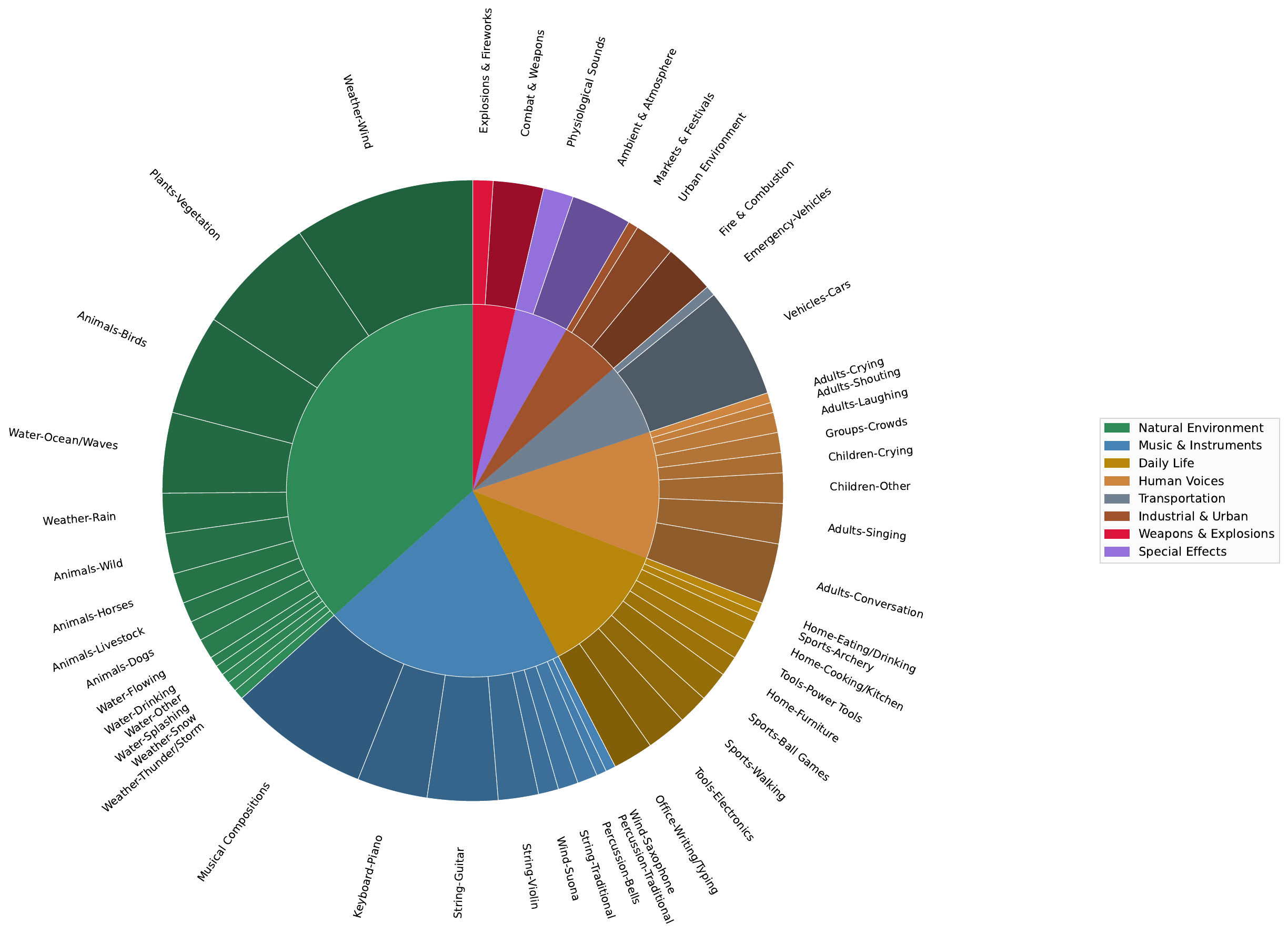}
    \caption{Statistical results of set1.}
    \label{fig:appendix_unibench_statistics_set1}
\end{figure*}

\begin{figure*}[h!]
    \centering
    \includegraphics[width=1.0\textwidth]{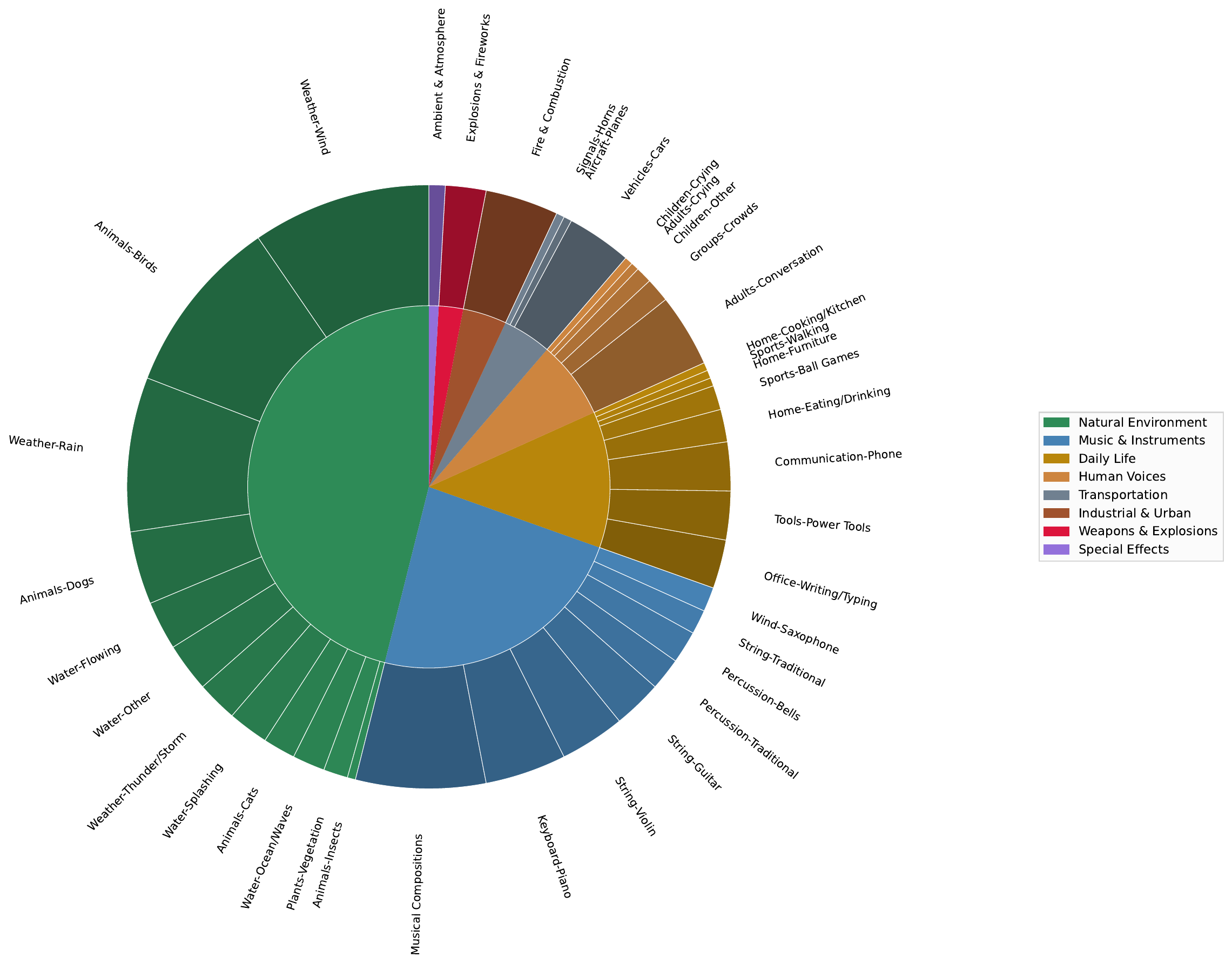}
    \caption{Statistical results of set2.}
    \label{fig:appendix_unibench_statistics_set2}
\end{figure*}

This section provides the detailed category catalog for Verse-Bench, along with a high-resolution pie chart illustrating its statistical distribution in Fig.~\ref{fig:appendix_unibench_statistics_all}~\ref{fig:appendix_unibench_statistics_set1}~\ref{fig:appendix_unibench_statistics_set2}.

Category list is in Tab.~\ref{tab:audio_stats_set1},~\ref{tab:audio_stats_set2},~\ref{tab:audio_stats_combined}.

\begin{table}[h!]
    \centering
    \caption{Detailed Audio Classification Statistics for Set 1}
    \label{tab:audio_stats_set1}
    \begin{tabularx}{\linewidth}{l S[table-format=2.0] S[table-format=2.1]}
        \toprule
        \textbf{Category / Subcategory} & \textbf{Count} & {\textbf{Percentage (\%)}} \\
        \midrule
        
        \textbf{Natural Environment} & \textbf{70} & \textbf{36.3} \\
        \quad Weather-Wind & 18 & {} \\
        \quad Plants-Vegetation & 12 & {} \\
        \quad Animals-Birds & 10 & {} \\
        \quad Water-Ocean/Waves & 8 & {} \\
        \quad \textit{Other Natural Subcategories} & 22 & {} \\ 
        
        \addlinespace 
        
        \textbf{Music \& Instruments} & \textbf{40} & \textbf{20.7} \\
        \quad Musical Compositions & 14 & {} \\
        \quad Keyboard-Piano & 7 & {} \\
        \quad String-Guitar & 7 & {} \\
        \quad \textit{Other Music Subcategories} & 12 & {} \\
        
        \addlinespace
        
        \textbf{Daily Life} & \textbf{22} & \textbf{11.4} \\
        \quad Office-Writing/Typing & 4 & {} \\
        \quad Tools-Electronics & 4 & {} \\
        \quad \textit{Other Daily Life Subcategories} & 14 & {} \\
        
        \addlinespace
        
        \textbf{Human Voices} & \textbf{21} & \textbf{10.9} \\
        \quad Adults-Conversation & 6 & {} \\
        \quad \textit{Other Human Voices Subcategories} & 15 & {} \\
        
        \addlinespace
        
        \textbf{Transportation} & \textbf{12} & \textbf{6.2} \\
        
        \addlinespace
        
        \textbf{Industrial \& Urban} & \textbf{10} & \textbf{5.2} \\
        
        \addlinespace
        
        \textbf{Special Effects} & \textbf{9} & \textbf{4.7} \\
        
        \addlinespace
        
        \textbf{Weapons \& Explosions} & \textbf{7} & \textbf{3.6} \\

        \midrule
        
        \textbf{Total Classified} & \textbf{191} & \textbf{93.2} \\
        Unclassified & 14 & 6.8 \\
        \midrule
        \textbf{Grand Total} & \textbf{205} & \textbf{100.0} \\
        
        \bottomrule
    \end{tabularx}
\end{table}

\begin{table}[h!]
    \centering
    \caption{Detailed Audio Classification Statistics for Set 2}
    \label{tab:audio_stats_set2}
    \begin{tabularx}{\linewidth}{l S[table-format=3.0] S[table-format=2.1]}
        \toprule
        \textbf{Category / Subcategory} & \textbf{Count} & {\textbf{Percentage (\%)}} \\
        \midrule
        
        \textbf{Natural Environment} & \textbf{106} & \textbf{35.9} \\
        \quad Weather-Wind & 22 & {} \\
        \quad Animals-Birds & 22 & {} \\
        \quad Weather-Rain & 19 & {} \\
        \quad Animals-Dogs & 9 & {} \\
        \quad \textit{Other Natural Subcategories} & 34 & {} \\
        
        \addlinespace
        
        \textbf{Music \& Instruments} & \textbf{54} & \textbf{18.3} \\
        \quad Musical Compositions & 16 & {} \\
        \quad Keyboard-Piano & 10 & {} \\
        \quad String-Violin & 8 & {} \\
        \quad \textit{Other Music Subcategories} & 20 & {} \\
        
        \addlinespace
        
        \textbf{Daily Life} & \textbf{28} & \textbf{9.5} \\
        \quad Office-Writing/Typing & 6 & {} \\
        \quad Tools-Power Tools & 6 & {} \\
        \quad Communication-Phone & 6 & {} \\
        \quad \textit{Other Daily Life Subcategories} & 10 & {} \\
        
        \addlinespace
        
        \textbf{Human Voices} & \textbf{16} & \textbf{5.4} \\
        \quad Adults-Conversation & 9 & {} \\
        \quad \textit{Other Human Voices Subcategories} & 7 & {} \\
        
        \addlinespace
        
        \textbf{Transportation} & \textbf{10} & \textbf{3.4} \\
        
        \addlinespace
        
        \textbf{Industrial \& Urban} & \textbf{9} & \textbf{3.1} \\
        
        \addlinespace
        
        \textbf{Weapons \& Explosions} & \textbf{5} & \textbf{1.7} \\

        \addlinespace

        \textbf{Special Effects} & \textbf{2} & \textbf{0.7} \\

        \midrule
        
        \textbf{Total Classified} & \textbf{230} & \textbf{78.0} \\
        Unclassified & 65 & 22.0 \\
        \midrule
        \textbf{Grand Total} & \textbf{295} & \textbf{100.0} \\
        
        \bottomrule
    \end{tabularx}
\end{table}

\begin{table}[h!]
    \centering
    \caption{Combined Audio Classification Statistics for Set 1 \& Set 2}
    \label{tab:audio_stats_combined}
    \begin{tabularx}{\linewidth}{l S[table-format=3.0] S[table-format=2.1]}
        \toprule
        \textbf{Category / Subcategory} & \textbf{Count} & {\textbf{Percentage (\%)}} \\
        \midrule
        
        \textbf{Natural Environment} & \textbf{176} & \textbf{36.1} \\
        \quad Weather-Wind & 40 & {} \\
        \quad Animals-Birds & 32 & {} \\
        \quad Weather-Rain & 23 & {} \\
        \quad Plants-Vegetation & 15 & {} \\
        \quad \textit{Other Natural Subcategories} & 66 & {} \\
        
        \addlinespace
        
        \textbf{Music \& Instruments} & \textbf{94} & \textbf{19.3} \\
        \quad Musical Compositions & 30 & {} \\
        \quad Keyboard-Piano & 17 & {} \\
        \quad String-Guitar & 13 & {} \\
        \quad String-Violin & 12 & {} \\
        \quad \textit{Other Music Subcategories} & 22 & {} \\
        
        \addlinespace
        
        \textbf{Daily Life} & \textbf{50} & \textbf{10.2} \\
        \quad Office-Writing/Typing & 10 & {} \\
        \quad Tools-Power Tools & 8 & {} \\
        \quad \textit{Other Daily Life Subcategories} & 32 & {} \\
        
        \addlinespace
        
        \textbf{Human Voices} & \textbf{37} & \textbf{7.6} \\
        \quad Adults-Conversation & 15 & {} \\
        \quad \textit{Other Human Voices Subcategories} & 22 & {} \\
        
        \addlinespace
        
        
        \textbf{Transportation} & \textbf{22} & \textbf{4.5} \\
        \quad Vehicles-Cars & 19 & {} \\
        
        \addlinespace
        
        \textbf{Industrial \& Urban} & \textbf{19} & \textbf{3.9} \\
        \quad Fire \& Combustion & 14 & {} \\

        \addlinespace
        
        \textbf{Weapons \& Explosions} & \textbf{12} & \textbf{2.5} \\

        \addlinespace

        \textbf{Special Effects} & \textbf{11} & \textbf{2.3} \\

        \midrule
        
        \textbf{Total Classified} & \textbf{421} & \textbf{84.2} \\
        Unclassified & 79 & 15.8 \\
        \midrule
        \textbf{Grand Total} & \textbf{500} & \textbf{100.0} \\
        
        \bottomrule
    \end{tabularx}
\end{table}

%% file: neurips_2025.bbl
\begin{thebibliography}{10}

\bibitem{song2020score}
Yang Song, Jascha Sohl-Dickstein, Diederik~P Kingma, Abhishek Kumar, Stefano Ermon, and Ben Poole.
\newblock Score-based generative modeling through stochastic differential equations.
\newblock {\em arXiv preprint arXiv:2011.13456}, 2020.

\bibitem{song2020improved}
Yang Song and Stefano Ermon.
\newblock Improved techniques for training score-based generative models.
\newblock {\em Advances in neural information processing systems}, 33:12438--12448, 2020.

\bibitem{lipman2022flow}
Yaron Lipman, Ricky~TQ Chen, Heli Ben-Hamu, Maximilian Nickel, and Matt Le.
\newblock Flow matching for generative modeling.
\newblock {\em arXiv preprint arXiv:2210.02747}, 2022.

\bibitem{peebles2023scalable}
William Peebles and Saining Xie.
\newblock Scalable diffusion models with transformers.
\newblock In {\em Proceedings of the IEEE/CVF international conference on computer vision}, pages 4195--4205, 2023.

\bibitem{openaisora2024}
OpenAI.
\newblock Video generation models as world simulators, 2024.

\bibitem{kong2024hunyuanvideo}
Weijie Kong, Qi~Tian, Zijian Zhang, Rox Min, Zuozhuo Dai, Jin Zhou, Jiangfeng Xiong, Xin Li, Bo~Wu, Jianwei Zhang, et~al.
\newblock Hunyuanvideo: A systematic framework for large video generative models.
\newblock {\em arXiv preprint arXiv:2412.03603}, 2024.

\bibitem{yang2024cogvideox}
Zhuoyi Yang, Jiayan Teng, Wendi Zheng, Ming Ding, Shiyu Huang, Jiazheng Xu, Yuanming Yang, Wenyi Hong, Xiaohan Zhang, Guanyu Feng, et~al.
\newblock Cogvideox: Text-to-video diffusion models with an expert transformer.
\newblock {\em arXiv preprint arXiv:2408.06072}, 2024.

\bibitem{wan2025wan}
Team Wan, Ang Wang, Baole Ai, Bin Wen, Chaojie Mao, Chen-Wei Xie, Di~Chen, Feiwu Yu, Haiming Zhao, Jianxiao Yang, et~al.
\newblock Wan: Open and advanced large-scale video generative models.
\newblock {\em arXiv preprint arXiv:2503.20314}, 2025.

\bibitem{wang2023progressive}
Duomin Wang, Yu~Deng, Zixin Yin, Heung-Yeung Shum, and Baoyuan Wang.
\newblock Progressive disentangled representation learning for fine-grained controllable talking head synthesis.
\newblock In {\em Proceedings of the IEEE/CVF Conference on Computer Vision and Pattern Recognition}, pages 17979--17989, 2023.

\bibitem{yu2023talking}
Zhentao Yu, Zixin Yin, Deyu Zhou, Duomin Wang, Finn Wong, and Baoyuan Wang.
\newblock Talking head generation with probabilistic audio-to-visual diffusion priors.
\newblock In {\em Proceedings of the IEEE/CVF International Conference on Computer Vision}, pages 7645--7655, 2023.

\bibitem{wang2025fantasytalking}
Mengchao Wang, Qiang Wang, Fan Jiang, Yaqi Fan, Yunpeng Zhang, Yonggang Qi, Kun Zhao, and Mu~Xu.
\newblock Fantasytalking: Realistic talking portrait generation via coherent motion synthesis.
\newblock {\em arXiv preprint arXiv:2504.04842}, 2025.

\bibitem{luo2025dreamactor}
Yuxuan Luo, Zhengkun Rong, Lizhen Wang, Longhao Zhang, Tianshu Hu, and Yongming Zhu.
\newblock Dreamactor-m1: Holistic, expressive and robust human image animation with hybrid guidance.
\newblock {\em arXiv preprint arXiv:2504.01724}, 2025.

\bibitem{tan2024animate}
Shuai Tan, Biao Gong, Xiang Wang, Shiwei Zhang, Dandan Zheng, Ruobing Zheng, Kecheng Zheng, Jingdong Chen, and Ming Yang.
\newblock Animate-x: Universal character image animation with enhanced motion representation.
\newblock {\em arXiv preprint arXiv:2410.10306}, 2024.

\bibitem{chen2025hunyuanvideo}
Yi~Chen, Sen Liang, Zixiang Zhou, Ziyao Huang, Yifeng Ma, Junshu Tang, Qin Lin, Yuan Zhou, and Qinglin Lu.
\newblock Hunyuanvideo-avatar: High-fidelity audio-driven human animation for multiple characters.
\newblock {\em arXiv preprint arXiv:2505.20156}, 2025.

\bibitem{lin2025omnihuman}
Gaojie Lin, Jianwen Jiang, Jiaqi Yang, Zerong Zheng, and Chao Liang.
\newblock Omnihuman-1: Rethinking the scaling-up of one-stage conditioned human animation models.
\newblock {\em arXiv preprint arXiv:2502.01061}, 2025.

\bibitem{shan2025hunyuanvideo}
Sizhe Shan, Qiulin Li, Yutao Cui, Miles Yang, Yuehai Wang, Qun Yang, Jin Zhou, and Zhao Zhong.
\newblock Hunyuanvideo-foley: Multimodal diffusion with representation alignment for high-fidelity foley audio generation.
\newblock {\em arXiv preprint arXiv:2508.16930}, 2025.

\bibitem{veo32025}
Google DeepMind.
\newblock Veo 3.
\newblock {\em https://https://deepmind.google/models/veo/}, 2025.5.

\bibitem{gong2025ace}
Junmin Gong, Sean Zhao, Sen Wang, Shengyuan Xu, and Joe Guo.
\newblock Ace-step: A step towards music generation foundation model.
\newblock {\em arXiv preprint arXiv:2506.00045}, 2025.

\bibitem{hamming1952mathematical}
R~Hamming.
\newblock Mathematical methods in large-scale computing units.
\newblock {\em Math Rev}, 13(1):495, 1952.

\bibitem{guo2023animatediff}
Yuwei Guo, Ceyuan Yang, Anyi Rao, Zhengyang Liang, Yaohui Wang, Yu~Qiao, Maneesh Agrawala, Dahua Lin, and Bo~Dai.
\newblock Animatediff: Animate your personalized text-to-image diffusion models without specific tuning.
\newblock {\em arXiv preprint arXiv:2307.04725}, 2023.

\bibitem{ho2022video}
Jonathan Ho, Tim Salimans, Alexey Gritsenko, William Chan, Mohammad Norouzi, and David~J Fleet.
\newblock Video diffusion models.
\newblock {\em Advances in neural information processing systems}, 35:8633--8646, 2022.

\bibitem{blattmann2023stable}
Andreas Blattmann, Tim Dockhorn, Sumith Kulal, Daniel Mendelevitch, Maciej Kilian, Dominik Lorenz, Yam Levi, Zion English, Vikram Voleti, Adam Letts, et~al.
\newblock Stable video diffusion: Scaling latent video diffusion models to large datasets.
\newblock {\em arXiv preprint arXiv:2311.15127}, 2023.

\bibitem{ma2025step}
Guoqing Ma, Haoyang Huang, Kun Yan, Liangyu Chen, Nan Duan, Shengming Yin, Changyi Wan, Ranchen Ming, Xiaoniu Song, Xing Chen, et~al.
\newblock Step-video-t2v technical report: The practice, challenges, and future of video foundation model.
\newblock {\em arXiv preprint arXiv:2502.10248}, 2025.

\bibitem{kuaishou2024kling}
Kuaishou.
\newblock Kling ai.
\newblock {\em https://klingai.kuaishou.com/}, 2024.06.

\bibitem{gao2025seedance}
Yu~Gao, Haoyuan Guo, Tuyen Hoang, Weilin Huang, Lu~Jiang, Fangyuan Kong, Huixia Li, Jiashi Li, Liang Li, Xiaojie Li, et~al.
\newblock Seedance 1.0: Exploring the boundaries of video generation models.
\newblock {\em arXiv preprint arXiv:2506.09113}, 2025.

\bibitem{polyak2024movie}
Adam Polyak, Amit Zohar, Andrew Brown, Andros Tjandra, Animesh Sinha, Ann Lee, Apoorv Vyas, Bowen Shi, Chih-Yao Ma, Ching-Yao Chuang, et~al.
\newblock Movie gen: A cast of media foundation models.
\newblock {\em arXiv preprint arXiv:2410.13720}, 2024.

\bibitem{veo22025}
Google DeepMind.
\newblock Veo 2.
\newblock {\em https://deepmind.google/technologies/veo/veo-2/}, 2024.12.

\bibitem{rombach2022high}
Robin Rombach, Andreas Blattmann, Dominik Lorenz, Patrick Esser, and Bj{\"o}rn Ommer.
\newblock High-resolution image synthesis with latent diffusion models.
\newblock In {\em Proceedings of the IEEE/CVF conference on computer vision and pattern recognition}, pages 10684--10695, 2022.

\bibitem{evans2025stable}
Zach Evans, Julian~D Parker, CJ~Carr, Zack Zukowski, Josiah Taylor, and Jordi Pons.
\newblock Stable audio open.
\newblock In {\em ICASSP 2025-2025 IEEE International Conference on Acoustics, Speech and Signal Processing (ICASSP)}, pages 1--5. IEEE, 2025.

\bibitem{riffusion}
Seth Forsgren and Hayk Martiros.
\newblock Riffusion: Stable diffusion for real-time music generation.
\newblock {\em https://github.com/riffusion/riffusion}, 2022.

\bibitem{audioldm2-2024taslp}
Haohe Liu, Yi~Yuan, Xubo Liu, Xinhao Mei, Qiuqiang Kong, Qiao Tian, Yuping Wang, Wenwu Wang, Yuxuan Wang, and Mark~D. Plumbley.
\newblock Audioldm 2: Learning holistic audio generation with self-supervised pretraining.
\newblock {\em IEEE/ACM Transactions on Audio, Speech, and Language Processing}, 32:2871--2883, 2024.

\bibitem{liu2023audioldm}
Haohe Liu, Zehua Chen, Yi~Yuan, Xinhao Mei, Xubo Liu, Danilo Mandic, Wenwu Wang, and Mark~D Plumbley.
\newblock {AudioLDM}: Text-to-audio generation with latent diffusion models.
\newblock {\em Proceedings of the International Conference on Machine Learning}, pages 21450--21474, 2023.

\bibitem{ning2025diffrhythm}
Ziqian Ning, Huakang Chen, Yuepeng Jiang, Chunbo Hao, Guobin Ma, Shuai Wang, Jixun Yao, and Lei Xie.
\newblock Diffrhythm: Blazingly fast and embarrassingly simple end-to-end full-length song generation with latent diffusion.
\newblock {\em arXiv preprint arXiv:2503.01183}, 2025.

\bibitem{ruan2023mm}
Ludan Ruan, Yiyang Ma, Huan Yang, Huiguo He, Bei Liu, Jianlong Fu, Nicholas~Jing Yuan, Qin Jin, and Baining Guo.
\newblock Mm-diffusion: Learning multi-modal diffusion models for joint audio and video generation.
\newblock In {\em Proceedings of the IEEE/CVF Conference on Computer Vision and Pattern Recognition}, pages 10219--10228, 2023.

\bibitem{ishii2024simple}
Masato Ishii, Akio Hayakawa, Takashi Shibuya, and Yuki Mitsufuji.
\newblock A simple but strong baseline for sounding video generation: Effective adaptation of audio and video diffusion models for joint generation.
\newblock {\em arXiv preprint arXiv:2409.17550}, 2024.

\bibitem{hayakawa2024mmdisco}
Akio Hayakawa, Masato Ishii, Takashi Shibuya, and Yuki Mitsufuji.
\newblock Mmdisco: Multi-modal discriminator-guided cooperative diffusion for joint audio and video generation.
\newblock {\em arXiv preprint arXiv:2405.17842}, 2024.

\bibitem{ergasti2025r}
Alex Ergasti, Giuseppe~Gabriele Tarollo, Filippo Botti, Tomaso Fontanini, Claudio Ferrari, Massimo Bertozzi, and Andrea Prati.
\newblock R-flav: Rolling flow matching for infinite audio video generation.
\newblock {\em arXiv preprint arXiv:2503.08307}, 2025.

\bibitem{lee2022sound}
Seung~Hyun Lee, Gyeongrok Oh, Wonmin Byeon, Chanyoung Kim, Won~Jeong Ryoo, Sang~Ho Yoon, Hyunjun Cho, Jihyun Bae, Jinkyu Kim, and Sangpil Kim.
\newblock Sound-guided semantic video generation.
\newblock In {\em European Conference on Computer Vision}, pages 34--50. Springer, 2022.

\bibitem{li2021ai}
Ruilong Li, Shan Yang, David~A Ross, and Angjoo Kanazawa.
\newblock Ai choreographer: Music conditioned 3d dance generation with aist++.
\newblock In {\em Proceedings of the IEEE/CVF international conference on computer vision}, pages 13401--13412, 2021.

\bibitem{liu2024syncflow}
Haohe Liu, Gael~Le Lan, Xinhao Mei, Zhaoheng Ni, Anurag Kumar, Varun Nagaraja, Wenwu Wang, Mark~D Plumbley, Yangyang Shi, and Vikas Chandra.
\newblock Syncflow: Toward temporally aligned joint audio-video generation from text.
\newblock {\em arXiv preprint arXiv:2412.15220}, 2024.

\bibitem{zhao2025uniform}
Lei Zhao, Linfeng Feng, Dongxu Ge, Rujin Chen, Fangqiu Yi, Chi Zhang, Xiao-Lei Zhang, and Xuelong Li.
\newblock Uniform: A unified multi-task diffusion transformer for audio-video generation.
\newblock {\em arXiv preprint arXiv:2502.03897}, 2025.

\bibitem{chen2020vggsound}
Honglie Chen, Weidi Xie, Andrea Vedaldi, and Andrew Zisserman.
\newblock Vggsound: A large-scale audio-visual dataset.
\newblock In {\em ICASSP 2020-2020 IEEE International Conference on Acoustics, Speech and Signal Processing (ICASSP)}, pages 721--725. IEEE, 2020.

\bibitem{gemmeke2017audio}
Jort~F Gemmeke, Daniel~PW Ellis, Dylan Freedman, Aren Jansen, Wade Lawrence, R~Channing Moore, Manoj Plakal, and Marvin Ritter.
\newblock Audio set: An ontology and human-labeled dataset for audio events.
\newblock In {\em 2017 IEEE international conference on acoustics, speech and signal processing (ICASSP)}, pages 776--780. IEEE, 2017.

\bibitem{chung2023unimax}
Hyung~Won Chung, Noah Constant, Xavier Garcia, Adam Roberts, Yi~Tay, Sharan Narang, and Orhan Firat.
\newblock Unimax: Fairer and more effective language sampling for large-scale multilingual pretraining.
\newblock {\em arXiv preprint arXiv:2304.09151}, 2023.

\bibitem{chen2024deep}
Junyu Chen, Han Cai, Junsong Chen, Enze Xie, Shang Yang, Haotian Tang, Muyang Li, Yao Lu, and Song Han.
\newblock Deep compression autoencoder for efficient high-resolution diffusion models.
\newblock {\em arXiv preprint arXiv:2410.10733}, 2024.

\bibitem{liao2024fish}
Shijia Liao, Yuxuan Wang, Tianyu Li, Yifan Cheng, Ruoyi Zhang, Rongzhi Zhou, and Yijin Xing.
\newblock Fish-speech: Leveraging large language models for advanced multilingual text-to-speech synthesis.
\newblock {\em arXiv preprint arXiv:2411.01156}, 2024.

\bibitem{wu2023exploring}
Haoning Wu, Erli Zhang, Liang Liao, Chaofeng Chen, Jingwen Hou, Annan Wang, Wenxiu Sun, Qiong Yan, and Weisi Lin.
\newblock Exploring video quality assessment on user generated contents from aesthetic and technical perspectives.
\newblock In {\em Proceedings of the IEEE/CVF International Conference on Computer Vision}, pages 20144--20154, 2023.

\bibitem{radford2023robust}
Alec Radford, Jong~Wook Kim, Tao Xu, Greg Brockman, Christine McLeavey, and Ilya Sutskever.
\newblock Robust speech recognition via large-scale weak supervision.
\newblock In {\em International conference on machine learning}, pages 28492--28518. PMLR, 2023.

\bibitem{deng2020retinaface}
Jiankang Deng, Jia Guo, Evangelos Ververas, Irene Kotsia, and Stefanos Zafeiriou.
\newblock Retinaface: Single-shot multi-level face localisation in the wild.
\newblock In {\em Proceedings of the IEEE/CVF conference on computer vision and pattern recognition}, pages 5203--5212, 2020.

\bibitem{chung2016out}
Joon~Son Chung and Andrew Zisserman.
\newblock Out of time: automated lip sync in the wild.
\newblock In {\em Asian conference on computer vision}, pages 251--263. Springer, 2016.

\bibitem{xu2025qwen2}
Jin Xu, Zhifang Guo, Jinzheng He, Hangrui Hu, Ting He, Shuai Bai, Keqin Chen, Jialin Wang, Yang Fan, Kai Dang, et~al.
\newblock Qwen2. 5-omni technical report.
\newblock {\em arXiv preprint arXiv:2503.20215}, 2025.

\bibitem{li2023mert}
Yizhi Li, Ruibin Yuan, Ge~Zhang, Yinghao Ma, Xingran Chen, Hanzhi Yin, Chenghao Xiao, Chenghua Lin, Anton Ragni, Emmanouil Benetos, et~al.
\newblock Mert: Acoustic music understanding model with large-scale self-supervised training.
\newblock {\em arXiv preprint arXiv:2306.00107}, 2023.

\bibitem{boito2024mhubert}
Marcely~Zanon Boito, Vivek Iyer, Nikolaos Lagos, Laurent Besacier, and Ioan Calapodescu.
\newblock mhubert-147: A compact multilingual hubert model.
\newblock {\em arXiv preprint arXiv:2406.06371}, 2024.

\bibitem{du2024cosyvoice}
Zhihao Du, Qian Chen, Shiliang Zhang, Kai Hu, Heng Lu, Yexin Yang, Hangrui Hu, Siqi Zheng, Yue Gu, Ziyang Ma, et~al.
\newblock Cosyvoice: A scalable multilingual zero-shot text-to-speech synthesizer based on supervised semantic tokens.
\newblock {\em arXiv preprint arXiv:2407.05407}, 2024.

\bibitem{du2024cosyvoice2}
Zhihao Du, Yuxuan Wang, Qian Chen, Xian Shi, Xiang Lv, Tianyu Zhao, Zhifu Gao, Yexin Yang, Changfeng Gao, Hui Wang, et~al.
\newblock Cosyvoice 2: Scalable streaming speech synthesis with large language models.
\newblock {\em arXiv preprint arXiv:2412.10117}, 2024.

\bibitem{peng2025vibevoice}
Zhiliang Peng, Jianwei Yu, Wenhui Wang, Yaoyao Chang, Yutao Sun, Li~Dong, Yi~Zhu, Weijiang Xu, Hangbo Bao, Zehua Wang, et~al.
\newblock Vibevoice technical report.
\newblock {\em arXiv preprint arXiv:2508.19205}, 2025.

\bibitem{gao2025wan}
Xin Gao, Li~Hu, Siqi Hu, Mingyang Huang, Chaonan Ji, Dechao Meng, Jinwei Qi, Penchong Qiao, Zhen Shen, Yafei Song, et~al.
\newblock Wan-s2v: Audio-driven cinematic video generation.
\newblock {\em arXiv preprint arXiv:2508.18621}, 2025.

\bibitem{jimeng}
ByteDance.
\newblock Jimeng ai, 2025.

\bibitem{wu2025qwen}
Chenfei Wu, Jiahao Li, Jingren Zhou, Junyang Lin, Kaiyuan Gao, Kun Yan, Sheng-ming Yin, Shuai Bai, Xiao Xu, Yilei Chen, et~al.
\newblock Qwen-image technical report.
\newblock {\em arXiv preprint arXiv:2508.02324}, 2025.

\bibitem{teed2020raft}
Zachary Teed and Jia Deng.
\newblock Raft: Recurrent all-pairs field transforms for optical flow.
\newblock In {\em European conference on computer vision}, pages 402--419. Springer, 2020.

\bibitem{yang2022maniqa}
Sidi Yang, Tianhe Wu, Shuwei Shi, Shanshan Lao, Yuan Gong, Mingdeng Cao, Jiahao Wang, and Yujiu Yang.
\newblock Maniqa: Multi-dimension attention network for no-reference image quality assessment.
\newblock In {\em Proceedings of the IEEE/CVF conference on computer vision and pattern recognition}, pages 1191--1200, 2022.

\bibitem{aesthetic-predictor-v2-5}
discus0434.
\newblock aesthetic-predictor-v2-5, 2024.

\bibitem{ke2021musiq}
Junjie Ke, Qifei Wang, Yilin Wang, Peyman Milanfar, and Feng Yang.
\newblock Musiq: Multi-scale image quality transformer.
\newblock In {\em Proceedings of the IEEE/CVF international conference on computer vision}, pages 5148--5157, 2021.

\bibitem{simeoni2025dinov3}
Oriane Sim{\'e}oni, Huy~V Vo, Maximilian Seitzer, Federico Baldassarre, Maxime Oquab, Cijo Jose, Vasil Khalidov, Marc Szafraniec, Seungeun Yi, Micha{\"e}l Ramamonjisoa, et~al.
\newblock Dinov3.
\newblock {\em arXiv preprint arXiv:2508.10104}, 2025.

\bibitem{kong2020panns}
Qiuqiang Kong, Yin Cao, Turab Iqbal, Yuxuan Wang, Wenwu Wang, and Mark~D Plumbley.
\newblock Panns: Large-scale pretrained audio neural networks for audio pattern recognition.
\newblock {\em IEEE/ACM Transactions on Audio, Speech, and Language Processing}, 28:2880--2894, 2020.

\bibitem{koutini2021efficient}
Khaled Koutini, Jan Schl{\"u}ter, Hamid Eghbal-Zadeh, and Gerhard Widmer.
\newblock Efficient training of audio transformers with patchout.
\newblock {\em arXiv preprint arXiv:2110.05069}, 2021.

\bibitem{wu2023large}
Yusong Wu, Ke~Chen, Tianyu Zhang, Yuchen Hui, Taylor Berg-Kirkpatrick, and Shlomo Dubnov.
\newblock Large-scale contrastive language-audio pretraining with feature fusion and keyword-to-caption augmentation.
\newblock In {\em ICASSP 2023-2023 IEEE International Conference on Acoustics, Speech and Signal Processing (ICASSP)}, pages 1--5. IEEE, 2023.

\bibitem{tjandra2025meta}
Andros Tjandra, Yi-Chiao Wu, Baishan Guo, John Hoffman, Brian Ellis, Apoorv Vyas, Bowen Shi, Sanyuan Chen, Matt Le, Nick Zacharov, et~al.
\newblock Meta audiobox aesthetics: Unified automatic quality assessment for speech, music, and sound.
\newblock {\em arXiv preprint arXiv:2502.05139}, 2025.

\bibitem{iashin2024synchformer}
Vladimir Iashin, Weidi Xie, Esa Rahtu, and Andrew Zisserman.
\newblock Synchformer: Efficient synchronization from sparse cues.
\newblock In {\em ICASSP 2024-2024 IEEE International Conference on Acoustics, Speech and Signal Processing (ICASSP)}, pages 5325--5329. IEEE, 2024.

\end{thebibliography}
